**Comparison of Waymo Rider-Only Crash Rates by Crash Type to Human Benchmarks at 56.7 Million Miles**

Kristofer D. Kusano[a], John M. Scanlon[a], Yin-Hsiu Chen[a], Timothy L. McMurry[a], Tilia Gode[a], Trent Victor[a]

[a] Waymo LLC, 1600 Amphitheatre Pkwy, Mountain View, CA 94043, United States

Corresponding author: Kristofer Kusano, kriskusano@waymo.com

**ABSTRACT**

**Objective:**

SAE Level 4 Automated Driving Systems (ADSs) are deployed on public roads, including Waymo's Rider-Only (RO) ride-hailing service (without a driver behind the steering wheel). The objective of this study was to perform a retrospective safety assessment of Waymo's RO crash rate compared to human benchmarks, including disaggregated by crash type.

**Methods:**

Eleven crash type groups were identified from commonly relied upon crash typologies that are derived from human crash databases. Human benchmarks were developed from state vehicle miles traveled (VMT) and police-reported crash data. Benchmarks were aligned to the same vehicle types, road types, and locations as where the Waymo Driver operated. Waymo crashes were extracted from the NHTSA Standing General Order (SGO). RO mileage was provided by the company via a public website. Any-injury-reported, Airbag Deployment, and Suspected Serious Injury+ crash outcomes were examined because they represented previously established, safety-relevant benchmarks where statistical testing could be performed at the current mileage.

**Results:**

Data was examined over 56.7 million RO miles through the end of January 2025, resulting in a statistically significant lower crashed vehicle rate for all crashes compared to the benchmarks in Any-Injury-Reported and Airbag Deployment, and Suspected Serious Injury+ crashes. Of the crash types, V2V Intersection crash events represented the largest total crash reduction, with a 96% reduction in Any-injury-reported (87%-99% confidence interval) and a 91% reduction in Airbag Deployment (76%-98% confidence interval) events. Cyclist, Motorcycle, Pedestrian, Secondary Crash, and Single Vehicle crashes were also statistically reduced for the Any-Injury-Reported outcome. There was no statistically significant disbenefit found in any of the 11 crash type groups.

**Conclusions:**

This study represents the first retrospective safety assessment of an RO ADS that made statistical conclusions about more serious crash outcomes (Airbag Deployment and Suspected Serious Injury+) and analyzed crash rates on a crash type basis. The crash type breakdown applied in the current analysis provides unique insight into the direction and magnitude of safety impact being achieved by a currently deployed ADS system. This work should be considered by stakeholders, regulators, and other ADS companies aiming to objectively evaluate the safety impact of ADS technology.

**INTRODUCTION**

Retrospective safety assessments, also known as safety impact or safety benefits studies, compare the in-field crash performance of vehicle safety technology to some benchmark. The approach, which often utilizes reported crash and exposure data, has been widely used to assess safety systems such as seatbelts (Elliot et al 2006, McCarthy 1989), airbags (McCartt and Kyrychenko 2007, Thompson et al 2002, Viano 2024), anti lock brakes (Kahane 1994), electronic stability control (Ferguson 2007), forward crash prevention (Cicchino 2017, Isaksson-Hellman and Lindman 2015, Fildes et al 2015), lane departure prevention (Cicchino 2018, Sternlund et al 2017), and other systems. The analytical approaches used in these retrospective safety assessments have included comparing crash rates (or insurance claims rates or amounts) before and after the system was deployed, as well as induced exposure (which allows for before and after comparisons without having an explicit exposure measure).

Many of the systems evaluated in the past are most effective at preventing or mitigating a certain type of crash. For example, forward crash prevention systems like Automatic Emergency Braking (AEB) and Forward Collision Warning (FCW) are most effective in front-to-rear (also known as rear-end) crashes. Therefore, many studies will present system effectiveness in reducing a target crash mode. The use of a conflict typology, also known as crash type groups, to subdivide the crash population to isolate common contributing factors and potential countermeasures has been a staple of traffic safety analysis (Kusano et al 2023, Najm and Smith 2007). Conflict typologies are particularly useful for analyzing safety performance with respect to the scenarios with the highest safety burden and among unique causal mechanisms (Kusano et al 2023).

Automated Driving Systems (ADSs), which are defined as SAE level 3 through 5 automation systems (SAE 2021), are just now becoming deployed on public roads in numbers that enable retrospective safety assessments. Past studies have quantified differences in ADS and human benchmark crash rates. Most of the literature on this subject has used early testing data from SAE level 4 ADS reported to the California Department of Motor Vehicle (DMV) and the National Highway Traffic Safety Administration (NHTSA) Standing General Order (SGO). This early testing data almost exclusively features an ADS with a human behind the steering wheel supervising the ADS with the capability to take over control of the vehicle if needed. See Scanlon et al (2024a) and Goodall (2021a, 2023) for thorough reviews of historical ADS safety impact studies to date. The presence of a human behind the wheel that can decide when to engage and disengage the ADS makes it difficult to separate the performance of the ADS from the human test drivers. Furthermore, the presence of a human supervising a level 4 system suggests a system under development that is not yet suitable for operation without a human supervisor. Schwall et al. (2020) found through a counterfactual simulation method that a human present was able to prevent 62% of crashes over 6.1 million miles of testing operations. This result highlighted the effect of a safety operator during early testing and the lack of comparability from testing operations to future on-road performance, unless counterfactual simulation is introduced after human disengagement (Schwall, et al., 2020). The most relevant performance of current level 4 ADS which are intended to be used as fleet-owned ride-hailing vehicles is when the ADS vehicle is operating without a human behind the wheel, or in rider-only (RO) configuration.

Because of the interest in ADS retrospective safety impact studies by many stakeholders, there have been recent efforts in developing best practices in performing and evaluating such studies. The RAVE Checklist, written by a group with safety impact research experience from automotive industry, insurance, and academia and described by Scanlon et al (2024b), is a list of requirements and recommendations to address the many challenges when performing retrospective safety impact studies. This study conforms to the required checklist items of the RAVE checklist (Scanlon et al 2024b, see appendix for RAVE checklist analysis for this study).

There are several recent studies that compared the aggregate, or overall, crash rate of a level 4 ADS in RO configuration to a human benchmark. Kusano et al (2024) compared Waymo's crash performance in several different crash outcomes to human benchmarks from Phoenix, San Francisco, and Los Angeles over 7 million RO miles. The study found a statistically significant reduction in police-reported (55% reduction) and any-injury-reported (80% reduction) crashes. No comparisons to higher severity outcomes, such as airbag deployments or serious injuries were done due to a lack of statistically significant conclusions due to limited RO miles relative to the benchmark rates. Chen and Shladover (2024) compared Waymo's RO crash rate to human benchmarks in San Francisco for just under 1 million RO miles at the any property damage or injury outcome level. No statistical testing was performed in that study, but the Waymo crash rate (reported as part of the NHTSA SGO) was found to be similar in magnitude to self-reported human transportation network company (TNC) crashes. It's unclear what definition of a crash is used for the self-reported TNC crash data, and whether that TNC crash definition is well matched to the ADS crashes reported as part of the NHTSA SGO. That is, there is an unknown amount of underreporting in the TNC crash data, while the ADS data from the SGO includes any amount of property damage with little to no underreporting. TNC drivers may have incentives to not report low severity collisions, as reported collisions may lead to deactivation from the platform. In two subsequent studies, Di Lillo et al (2024a, 2024b) compared Waymo's 3rd party liability property damage and bodily injury claims rate to a human benchmark weighted by garaged zip code proportional to the miles driven by the Waymo RO fleet over the first 3.9 million and 25.3 million miles. A 3rd party liability claim is when a party involved in a crash asks for payment from a party's insurance. These studies use a 3rd party liability claim that is paid as a proxy for whether the insured vehicle contributes to the crash, which is a complementary view of crash involvement to the overall crash rates analyses studied by Kusano et al (2024), which compares crash rates regardless of contribution to the crash. Using the 25.3 million mile analysis (superseding the 3.9 million mile analysis), Di Lillo et al (2024b) found a statistically significant reduction in both property damage (88% reduction) and bodily injury (92% reduction) claims. Di Lillo et al (2024b) also compared Waymo RO 3rd party liability property damage and bodily injury claims rates to a benchmark of human insured latest-generation vehicles (model years 2018-2021), which have a lower claims rate than all human driven vehicles. The study found Waymo RO had an 86% reduction in property damage claims and 90% reduction in bodily injury claims compared to the latest-generation human driven vehicle benchmark.

As ADS deployments have continued to operate and expand to collect additional miles, there is now an opportunity to do such a safety impact analysis of more rare safety outcomes (such as serious injuries) and to disaggregate analysis by crash type as has been done in the past for other vehicle safety systems. Both of these types of analyses require sufficient mileage for statistical comparison and have thus been limited in the past. As the benchmark becomes more rare (i.e., a lower crash rate), more miles and/or a larger relative difference in performance between the ADS and benchmark is needed to draw statistically significant conclusions. For example, Scanlon et al (2024a) performed an example statistical power analysis that computed the number of miles needed for statistical significance for hypothetical ADS with different performances relative to the benchmarks. An ADS with a crash rate of 10% the national suspected serious injury+ benchmark (i.e., a 90% reduction of the benchmark of 0.11 Incidents per Million Miles, IPMM) would require 56.3 million miles. Waymo's RO miles are now within this range where statistical conclusions could be drawn about such a serious injury. The primary focus of road safety, aligned with the Vision Zero movement, is to eliminate serious and fatal injuries (Lie and Tingvall 2024). Thus, retrospective evidence of the performance of level 4 ADS would serve as a continuous confidence growth in the safety assurance process (Favaro et al 2024) that used design-based and prospective studies during system development (for example, Scanlon et al 2021). Similarly, disaggregating the benchmark comparison by crash type reduces the benchmark crash rate under test, increasing the mileage requirement. Because current level 4 ADS are deployed where the system is responsible for the entire dynamic driving task without the ability for humans to take over at any time, the ADS may have a safety impact on all types of crashes. Until now, retrospective safety assessment studies of ADS in RO configuration have only compared aggregate crash rates, including all types of crashes, as that has been the only level of analysis possible. Analyses of performance by crash type may provide insight into how level 4 ADS are achieving reduced overall crash rates and what distributional shifts in crash type are occurring.

This study conducted a retrospective safety impact analysis of Waymo's level 4 ADS in RO configuration on surface streets in San Francisco, Phoenix, Los Angeles, and Austin over 56.7 million RO miles with two primary research questions: (1) what is Waymo's aggregate (all crash type) crash performance relative to aligned human benchmarks in the outcomes of *Any-Injury-Reported*, *Airbag Deployment*, and *Suspected Serious Injury+* outcomes? and (2) what is Waymo's crash rate for disaggregated crash types relative to human benchmarks in the *Any-Injury-Reported* and *Airbag Deployment* outcomes? This study has several contributions compared to past ADS safety impact studies. First, this study compares higher severity outcomes (*Airbag Deployment* and *Suspected Serious Injury+*) to human benchmarks than previous studies. Second, past studies have not compared RO ADS performance to these human benchmarks disaggregated by crash type. Third, this study applied a spatial dynamic benchmark correction described by Chen et al (2024) that adjusts the human benchmarks proportional to the spatial distribution of driving of the RO fleet within the geographies they operate in.

**METHODS**

**ADS and Benchmark Data Alignment**

The methodology was designed according to the RAVE checklist (Scanlon et al 2024b) and was implemented in four main steps that are shown in Figure 1. First, raw data was extracted from a variety of available data sources. Next the mileage and crash data from the benchmark and ADS crash data sources were aligned to maximize comparability. Specifically, the mileage and crash benchmark data was restricted to passenger vehicles and surface streets, and dynamic spatial adjustments were performed to make the benchmarking driving distribution representative of the ADS driving. Crash rates were generated by both crash types and various outcome levels. Statistical testing was then used to evaluate the meaningfulness of any observed differences in crash rates.

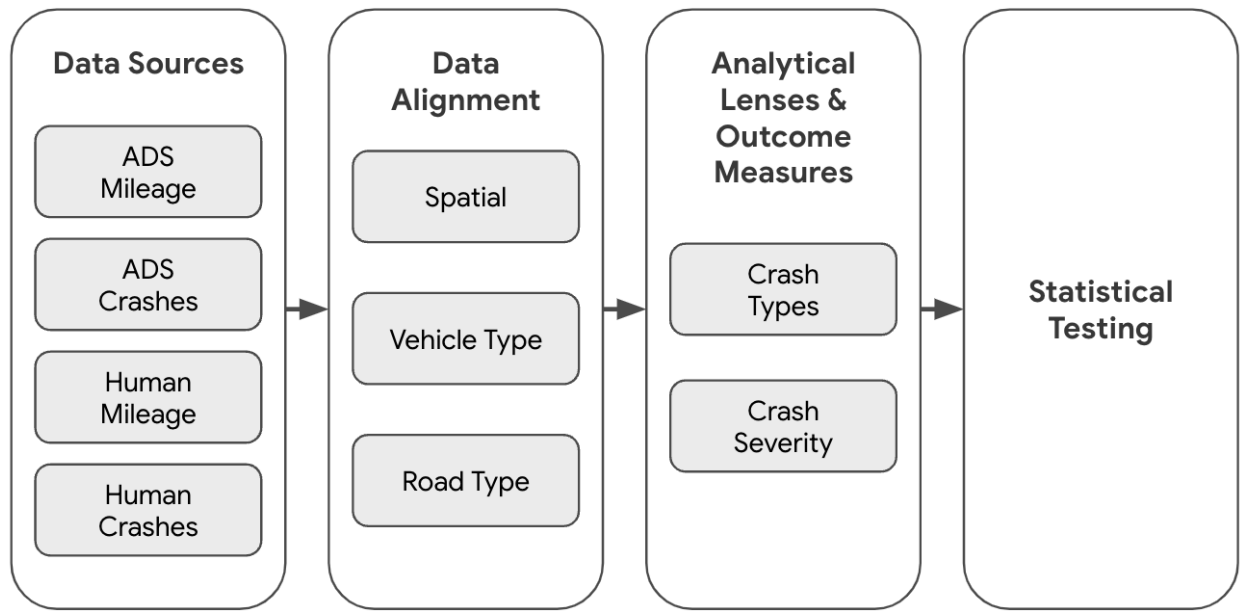

**Figure 1.** Data Process Methodology for ADS and Benchmark Comparison.

This study compared crash rates for Waymo's RO service derived from the NHTSA SGO and self-reported mileage by Waymo to human benchmarks derived from crash and Vehicle Miles Traveled (VMT) databases maintained by the states of California, Arizona, and Texas. A full listing of these databases relied upon can be found in Table 1. Details of the data sources are in the appendix.

**Table 1.** Data sources relied upon in the current study.

| Driver | Geographic Region | Crash Source | Mileage Source |
|---|---|---|---|
| Waymo ADS | Arizona, California, Texas | NHTSA SGO (NHTSA 2023) | Waymo Safety Impact Data Hub (Waymo 2025) |
| Benchmark | Arizona | ADOT (ADOT 2023a, 2023c) | AZ Certified Public Miles (ADOT 2023b) + FHWA (FHWA 2023a) |
| | California | CA SWITRS (CHP 2024) | CA Public Road Data (Caltrans 2023) + FHWA (FHWA 2023a) |
| | Texas | TXDOT Crash Records Information System (CRIS) (TXDOT 2022) | TX Transit Statistics (TXDOT 2023) + FHWA (FHWA 2023a) |

Data alignment promotes comparability between datasets. This is sometimes referred to colloquially as making an "apples-to-apples" comparison (Scanlon et al. 2024b). This study performed alignment along four main dimensions known to influence crash risk, including (1) vehicle type, (2) road type, and (3) spatial driving distribution. Vehicle type and road type were accounted for through subselection. A dynamic benchmarking routine previously developed by Chen et al (2024) was used to adjust for crash risk by where within their deployed geographic regions the ADS operated.

The Waymo RO operations in this study are identical to those described in Kusano et al (2024). Waymo's current RO operations have exclusively taken place using recent model run Chrysler Pacifica (not actively in operation) and Jaguar I-Pace (active) platforms, which are both classified as passenger vehicles according to the 49 CFR § 565.15 classification. For this study, surface streets refers to all roadways that are not "Interstates" or "Other Arterials - Other Freeways and Expressways" as defined using FHWA's highway function classification coding (FHWA 2023b). The existing RO operations mostly occurred on surface streets, so this study only examines miles and crashes that occurred on surface streets. A final dimension considered was "in-transport" status, which refers to all vehicles that are not in designated parking, parked off the roadway, parked on private property, or working vehicles. Mileage is only accumulated while "in-transport", and this variable is readily available in all police-reported databases, which makes it a straightforward dimension to align the data on. See the appendix for the procedure for determining in-transport status using SGO data, which is identical to the method used in Kusano et al (2024).

The benchmark mileage and crashes include data from multiple vehicle and road types, and quantifying in-transport surface street, passenger vehicle rates involves a combination of subselection, data joining, and re-weighting. This process was previously described in Scanlon et al. (2024a) with the current study extending the methodology to Texas data and developing individual benchmarks for 11 crash type groups. The crash data was directly subset for these three data requirements. The mileage data was subset for surface streets. The amount of mileage in each region was not broken down by vehicle type, so the total mileage (including all vehicle types) reported by the states was adjusted based on the proportion of passenger vehicle miles reported in the FHWA VM-4 tables. By definition, all mileage data is accumulated while "in-transit". Each mileage and crash dataset has unique features for identifying surface streets, passenger vehicles, and in-transport status. The variables relied upon in this study are shown in the appendix.

**Spatial Dynamic Benchmark**

The benchmark data relied upon were examined as four distinct geographic areas: Travis, Hays, and Williamson Counties in Texas (Austin); Los Angeles County, California; Maricopa County, Arizona (Phoenix); and San Francisco County, California. The dynamic benchmarking routine described presented by Chen et al (2024) and briefly summarized in the following paragraphs then effectively further subset and weighted the benchmark data to only include the area within the selected counties where the Waymo RO service drove, proportional to the miles driven.

Relying directly on the entirety of crash and mileage data from the counties making up the Waymo RO service area would effectively create what will be referred to as "unadjusted" benchmark crash rates, whereby the crash rates are representative of where the current driving population currently aggregates VMT. Waymo's ODD has gradually changed over time and does not necessarily include the entire counties. Additionally, Waymo operates as a ride-hailing fleet with unique driving patterns that are responsive to user demand. Because of this, Chen et al. (2024) examined the spatial distribution of where VMT was being accumulated and noted distinct differences in the driving distributions. The driving mix differences had direct implications on the crash risk.

Chen et al. (2024) created a "dynamic" benchmark routine reweighting the human benchmark data to reflect the driving distribution of the ADS systems. This effectively models the crash rate of the benchmark given that the benchmark population drove with the same spatial distribution as the Waymo driver. For a spatial adjustment, the routine discretizes the miles driven by the ADS (Waymo) and benchmark (HPMS human driven mileage) into level-13 S2 cells thus providing a spatial distribution of driving miles throughout some bounded geographic area. The proportion of driving miles driven by both the Waymo and human within a given cell is used to reweight benchmark data. Likewise, if an area of the benchmark data geographic region is not driven in by the Waymo Driver, it is re-weighted in a manner that essentially excludes it from the derived benchmark (zero-weighted).

**Crash Type Analysis**

One of the contributions of this paper is to compare crash rates in individual crash types. There are many frameworks to consider for differentiating between crash types (Najm and Smith 2007, Kusano et al 2023). There are two competing priorities that need to be considered. First, which crash types are informative for evaluating safety performance? Second, is there enough driving exposure to identify statistical differences? There is an inherent tradeoff in selecting crash type groupings, where a crash type grouping with too many categories could draw more specific conclusions about ADS performance but would suffer from a lack of statistical power. A crash type grouping with too few categories would have higher statistical power, but not aid in the understanding of ADS performance in specific crash types.

To balance the tradeoff of crash type groupings between analysis ability and statistical power, the approach in this study was to consider two dimensions, crash partner type and geometric configuration, in deriving crash type groupings. Crash types were only assigned for the first two involved parties (or one vehicle if it was a single vehicle crash). Vehicles involved in secondary contact events were indicated accordingly. Figure 2 shows the crash type groupings used in this study. Altogether, these crash types encompass 88% of the total police-reported crashes with at least minor injuries and 86% of the total fatal collisions nationally in the US (Kusano et al 2023). Generally speaking, a crash partner type and geometrical lens is informative about (a) avoidability and (b) potential severity, and tend to share similar sets of causal mechanisms (Kusano et al 2023).

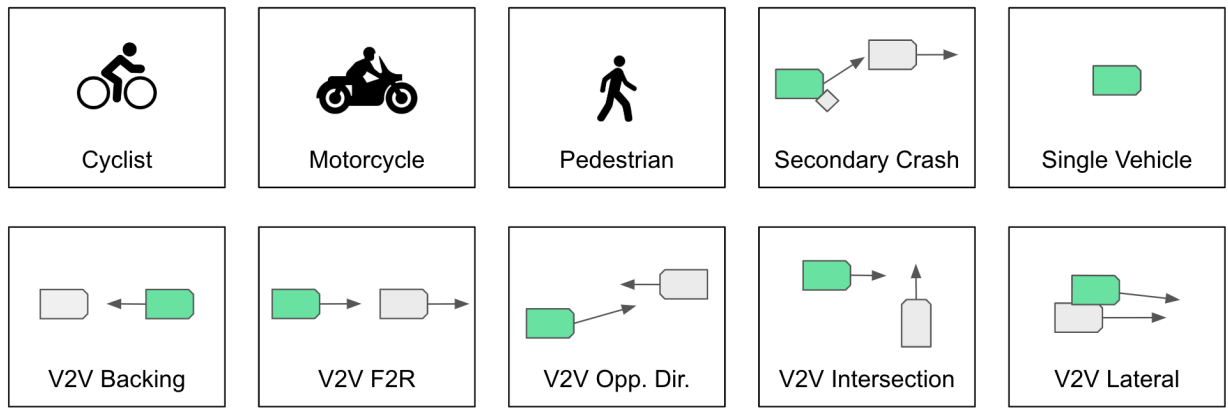

**Figure 2.** Crash Type Groupings for ADS and Human Benchmark Crashed Vehicle Rate Comparisons. The abbreviation "V2V" stands for Vehicle-to-Vehicle. The abbreviation F2R stands for Front-to-Rear. An 11th group, "All Others" is not pictured.

Cyclist, motorcycle, and pedestrian (often referred to as vulnerable road user or VRU) crashes were each examined as individual groups because passenger vehicle collisions with VRUs in the outcome groups examined in this study are rare, which would not support splitting these crashes into more groups based on maneuvers at the current VMT. Next, various vehicle-to-vehicle (V2V) crash groupings were examined: backing, front-to-rear (F2R), opposite direction (Opp. Dir.), intersection, and lateral. Single vehicle crashes (involving the passenger vehicle striking an object or the ground) were also examined as their own group. Secondary crashes, where a vehicle is involved in a crash with another vehicle that had previously crashed, were separated as their own group. All other crashes were classified into an other crash category, which includes missing or unknown crash types in the human benchmark data. These groups were chosen based on the highest level aggregations of crashes used by NHTSA in their standardized crash databases (e.g., the Crash Reporting Sampling System) and other typologies (Najm and Smith 2007, Kusano et al 2023). In total, there were 11 crash type groups examined in this study.

**Outcome Levels**

A number of outcome levels were considered for the current analysis based on previously established levels and what was readily possible from the underlying data sources. Scanlon et al. (2024a) outlined multiple severity levels potentially useful in a benchmarking analysis, which range from any amount of property damage to fatal injuries. Consistent with the analysis and recommendations provided by Scanlon et al. (2024a), outcome levels were pre-selected to minimize potential bias in reporting between the benchmark and ADS population. Using police-reported data as a benchmark, there are underreporting considerations and geographic-specific reporting thresholds. It is difficult to draw conclusions about an *Any Property Damage or Injury* outcome level, which included all in-transport and impacted ADS crashes, because the human data has uncertainty in the lower reporting threshold and underreporting (Kusano et al. 2024). Additionally, because some reportable crashes are not reported to or by police, and the degree to which this underreporting occurred is unknown, a police-reported threshold was not considered. As discussed in Scanlon et al. (2024), the *Police-Reported* human benchmarks may suffer from systematic underreporting, especially in California, where the state police report crash database does not require police jurisdictions to report property damage only crashes. Scanlon et al. (2024) also noted challenges with a *tow-away* outcome level and recommended against its usage. ADS-equipped vehicles can be towed for a variety of reasons during only minor damage collisions, which makes comparability to human-driven vehicles challenging. For completeness, the comparison of ADS and benchmark crash rates for the *Police-Reported* and *Any Property Damage or Injury* outcome are provided with the online supplemental materials, but for the aforementioned reasons, these estimates are considered less credible and are not a focus of the current study.

As the traditional focus of traffic safety research has been on preventing serious and fatal injuries, this study examined *Any-Injury-Reported*, *Airbag Deployment*, and *Suspected Serious Injury+* outcomes. These outcome levels are the most injury-relevant outcomes that are readily available in both human and ADS crash data and where there is sufficient ADS mileage to draw statistically relevant conclusions.

The NHTSA SGO and benchmark crash data was subset by observed outcomes in order to align the comparisons. SGO crashes were limited to those where the ADS vehicle was in-transport (i.e., not parked in a parking space) and impacted (i.e., the Waymo ADS vehicle striked or was struck by another road user or object). The *Suspected Serious Injury+*, *Airbag Deployment*, and *Any-Injury-Reported* outcomes were the primary focus of this study. A *Suspected Serious Injury+* is a crash where someone involved sustains a "Killed" or "Incapacitating" police-reported injury. For example, "A" injuries in California police reports on the KABCO scale are "severe laceration resulting in exposure of underlying tissues/muscles/organs or resulting in significant loss of blood", "broken or distorted extremity (arm or leg)", "crush injuries, "suspected skull, chest or abdominal injuries other than bruises or lacerations," "Significant burns (second and third degree burns over 10% or more of the body)", "unconsciousness when taken from the collision scene", and/or "paralysis" (CHP, 2017). Police report data was obtained through public record requests for the three SGO crashes with "Serious" or "Fatality" SGO maximum severity. A police report obtained for case 30270-8968 indicated the sole injury reported in the crash was reported as a "complain of pain" (or "C" injury) on the police report, but was reported as a "Serious" injury in the SGO because that occupant was transported in an ambulance to seek medical treatment. The other two "Serious" or "Fatality" SGO severity crashes indicated an "Incapacitating" or "Killed" maximum severity on the police reports. As the human benchmarks use the police-reported injury severity, only the two (2) SGO-reported crashes with confirmed "K" or "A" maximum severity from police reports were included in the *Suspected Serious Injury+* crashes. Both *Suspected Serious Injury+* crashes involving a Waymo vehicle during the study period were Secondary Crashes, meaning the Waymo was not involved in the first event in the crash sequence. See the appendix for a complete description of the *Suspected Serious Injury+* crashes.

An *Airbag Deployment* crash is where one or more vehicles involved deploys any airbag due to the crash. An *Any-Injury-Reported* crash is where any level of injury is reported due to the crash. Note that outcomes were classified at the crash level including all parties involved in the crash. The injury outcome levels (*Suspected Serious Injury+* and *Any-Injury-Reported*) were selected if any party in the crash was injured, whether riding in the Waymo vehicle or otherwise. Similarly, the *Airbag Deployment* outcome was selected if any vehicle involved in the collision sequence had an airbag deploy, not just the Waymo vehicle. The classification routines and variables used for each benchmark dataset can be found in the appendix. The *Any-Injury-Reported* benchmark utilized the same underreporting correction as described in Scanlon et al (2024a). No underreporting correction was applied to the *Airbag Deployment* and *Suspected Serious Injury+* benchmarks, as no data is available to estimate the amount of underreporting in these outcome levels. There is reason to believe that the underreporting in human crashes in these outcomes are non-zero.

**Statistical Testing**

A statistical comparison between the ADS and benchmark crash rates was done using Clopper-Pearson limits to estimate 95% confidence intervals for the ratio of two Poisson mean occurrence rates, as described by Nelson (1970, Appendix I), which is the same method adopted by Kusano et al (2024).

**RESULTS**

**ADS and Benchmark Population Comparisons**

The ADS and human benchmark data were extracted from the same locations (counties), vehicle types (passenger vehicles), and road types (surface streets) in an attempt to account for these factors that can affect crash rates. The dynamic benchmark adjustment for spatial driving mix further aligns the benchmark and ADS driving population. The research question of this study addresses a comparison between the Waymo RO service and the overall driving population in these areas. Therefore, human driver demographics are not of particular importance because the entire driving population is represented in the data. Table 2 shows the number of Waymo RO miles by location and calendar year. Most of the benchmark data is from calendar year 2022, which is the last year with complete data available at the time of writing.

**Table 2.** Waymo Rider-Only Millions of Miles by Calendar Year and Location.

| Location | 2019 | 2020 | 2021 | 2022 | 2023 | 2024 | Jan 2025 |
|---|---|---|---|---|---|---|---|
| Phoenix, AZ | 0.0205 | 0.0739 | 0.328 | 0.454 | 6.52 | 20.9 | 2.83 |
| San Francisco, CA | 0 | 0 | 0 | 0.0700 | 2.46 | 13.5 | 2.23 |
| Los Angeles, CA | 0 | 0 | 0 | 0 | 0.140 | 5.03 | 1.28 |
| Austin, TX | 0 | 0 | 0 | 0 | 0 | 0.555 | 0.278 |
| Mountain View, CA | 0 | 0 | 0 | 0 | 0 | 0.005 | 0.003 |
| Atlanta, GA | 0 | 0 | 0 | 0 | 0 | 0 | 0.0002 |

**Aggregate Crash Rate Comparison**

Table 3 lists the number of events with the *Any-Injury-Reported*, *Airbag Deployment*, and *Suspected Serious Injury+* outcomes by location for the Waymo RO service during the study period. During the study period, there were 31.159 million miles in Phoenix, 18.260 million miles in San Francisco, 6.448 million miles in Los Angeles, and 0.834 million miles in Austin driven by the Waymo RO service for a total mileage of 56.700 million miles. Table 4 shows the comparison of the aggregate Waymo RO and benchmark crash rates for the *Any-Injury-Reported*, *Airbag Deployment*, and *Suspected Serious Injury+* outcomes. These comparisons were not statistically significant for Los Angeles due to limited mileage. The point estimates in Los Angeles, however, are of similar magnitudes than those in Phoenix and San Francisco. The Waymo RO service had a statistically significant reduction in *Any-Injury-Reported* and *Airbag Deployment* outcomes in Phoenix, San Francisco, and all locations combined. The Waymo RO service had a statistically significant reduction in *Suspected Serious Injury+* crashes when considering all locations combined, with a 85% reduction (39% to 99% reduction 95% confidence interval), in Phoenix, with a 100% reduction (3% to 100% reduction 95% confidence interval), and in San Francisco with a 76% reduction (3% to 98% reduction 95% confidence interval).

**Table 3.** Event Counts by Outcome and Location (through January 2025, 56.7M RO Miles).

| Location | Any-Injury-Reported | Airbag Deployment | Suspected Serious Injury+ |
|---|---|---|---|
| Phoenix | 24 | 8 | 0 |
| San Francisco | 16 | 7 | 2 |
| Los Angeles | 8 | 2 | 0 |
| Austin | 0 | 1 | 0 |
| All Locations | 48 | 18 | 2 |

**Table 4.** Comparison of Waymo RO and Human Benchmark Crashed Vehicle Rates for *Any-Injury-Reported*, *Airbag Deployment*, and *Suspected Serious Injury+* Crashes (through January 2025, 56.7M RO Miles).

| Outcome | Benchmark | Location | Human IPMM | ADS IPMM | Expected ADS Count Different to Benchmark | ADS to Human Percent Difference | 95% Confidence Interval | |
|---|---|---|---|---|---|---|---|---|
| *Any-Injury-Reported* | Scanlon et al. (2024a) - Blincoe-Adjusted with Dynamic Adjustment | Phoenix | 2.09 | 0.77 | -41.1 | **-63%\*** | **-78%** | **-42%** |
| | | San Francisco | 8.02 | 0.88 | -130.4 | **-89%\*** | **-94%** | **-81%** |
| | | All Locations - Mileage Blended | 4.04 | 0.85 | -181.2 | **-79%\*** | **-85%** | **-71%** |
| *Airbag Deployment* | Scanlon et al. (2024a) - Observed with Dynamic Adjustment | Phoenix | 1.42 | 0.26 | -36.3 | **-82%\*** | **-93%** | **-62%** |
| | | San Francisco | 2.31 | 0.38 | -35.1 | **-83%\*** | **-94%** | **-63%** |
| | | All Locations - Mileage Blended | 1.69 | 0.32 | -77.9 | **-81%\*** | **-90%** | **-69%** |
| *Suspected Serious Injury+* | Scanlon et al. (2024a) - Observed with Dynamic Adjustment | Phoenix | 0.12 | 0.00 | -3.8 | **-100%\*** | **-100%** | **-3%** |
| | | San Francisco | 0.46 | 0.11 | -6.5 | **-76%\*** | **-98%** | **-3%** |
| | | All Locations - Mileage Blended | 0.24 | 0.04 | -11.3 | **-85%\*** | **-99%** | **-39%** |

**Crash Rate Comparison by Crash Type**

The number of observed Waymo events by outcome level are listed in the appendix. Table 5 compares the Waymo RO and benchmark crash rates by crash type for the *Any-Injury-Reported* outcome in all locations combined. Results for individual locations by crash type are included in the appendix. The results show a statistically significant reduction in *Any-Injury-Reported* crashes in Cyclist, Motorcyclist, Pedestrian, Secondary Crash, Single Vehicle, V2V Intersection, and V2V Lateral crash types in all locations combined. When evaluating individual locations, the same crash types had a statistically significant reduction in *Any-Injury-Reported* crashes in San Francisco except for Secondary Crashes. In addition, there was a statistically significant reduction in V2V F2R crashes in San Francisco. Only V2V Intersections crashes had a statistically significant reduction for *Any-Injury-Reported* crashes in Phoenix. The reduction in V2V Intersections was also statistically significant in Los Angeles.

**Table 5.** Comparison of ADS and Human Benchmark (with Dynamic Benchmark Adjustment) Crashed Vehicle Rates in All Locations Combined by Crash Type in *Any-Injury-Reported* Crashes (through January 2025, 56.7M RO Miles).

| Crash Type | Human IPMM | ADS IPMM | Expected ADS Count Different to Benchmark | ADS to Human Percent Difference | 95% Confidence Interval | |
|---|---|---|---|---|---|---|
| Cyclist | 0.29 | 0.05 | -13.5 | **-82%*** | **-97%** | **-41%** |
| Motorcycle | 0.20 | 0.04 | -9.4 | **-82%*** | **-99%** | **-29%** |
| Pedestrian | 0.42 | 0.04 | -21.8 | **-92%*** | **-99%** | **-66%** |
| Secondary Crash | 0.21 | 0.07 | -7.9 | **-66%*** | **-93%** | **-5%** |
| Single Vehicle | 0.27 | 0.02 | -14.2 | **-93%*** | **-100%** | **-58%** |
| V2V Backing | 0.04 | 0.00 | -2.0 | -100% | -100% | 87% |
| V2V F2R | 0.61 | 0.44 | -9.5 | -28% | -56% | 12% |
| V2V Opposite Direction | 0.06 | 0.04 | -1.6 | -45% | -95% | 126% |
| V2V Intersection | 1.58 | 0.07 | -85.7 | **-96%*** | **-99%** | **-87%** |
| V2V Lateral | 0.27 | 0.07 | -11.5 | **-74%*** | **-94%** | **-27%** |
| Other | 0.09 | 0.02 | -3.9 | -80% | -100% | 31% |

Table 6 compares the Waymo RO and benchmark crash rates by crash type for the *Airbag Deployment* outcome in all locations combined. Results for individual locations by crash type are included in the appendix. In all locations, the Single Vehicle and V2V Intersection crash type had a statistically significant reduction in *Airbag Deployment* crashes. The reduction in *Airbag Deployment* V2V Intersection crash type was also statistically significant when considering Phoenix, San Francisco, and Los Angeles alone. Additionally, there was a statistically significant reduction in *Airbag Deployment* Single Vehicle crashes in San Francisco and Secondary Crashes in Phoenix. Other crash type comparisons were not statistically significant. See the appendix for a description of additional trends in select crash types.

**Table 6.** Comparison of ADS and Human Benchmark (with Dynamic Benchmark Adjustment) Crashed Vehicle Rates in All Locations Combined by Crash Type in *Airbag Deployment* Crashes (through January 2025, 56.7M RO Miles).

| Crash Type | Human IPMM | ADS IPMM | Expected ADS Count Different to Benchmark | ADS to Human Percent Difference | 95% Confidence Interval | |
|---|---|---|---|---|---|---|
| Secondary Crash | 0.12 | 0.04 | -4.6 | -70% | -97% | 23% |
| Single Vehicle | 0.19 | 0.00 | -10.9 | **-100%*** | **-100%** | **-66%** |
| V2V Backing | 0.01 | 0.00 | -0.3 | -100% | -100% | 1001% |
| V2V F2R | 0.22 | 0.12 | -5.3 | -43% | -80% | 27% |
| V2V Opposite Direction | 0.05 | 0.04 | -0.6 | -24% | -94% | 209% |
| V2V Intersection | 0.96 | 0.09 | -49.2 | **-91%*** | **-98%** | **-76%** |
| V2V Lateral | 0.09 | 0.02 | -4.3 | -81% | -100% | 20% |
| Other | 0.04 | 0.02 | -1.3 | -57% | -99% | 173% |

Figure 3 and Figure 4 compare the number of crashes a driver with the average human benchmark rate driving the same distance as the Waymo RO service and the experienced number of Waymo crashes along with percent reductions.

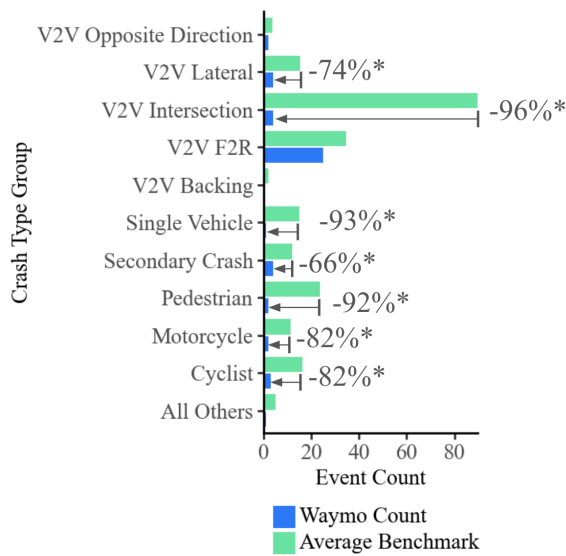

**Figure 3.** Comparison of Observed Waymo and Average Benchmark *Any-Injury-Reported* Crashes in All Locations over 56.7 Million Miles. Comparisons labeled with an asterisk (*) had a statistically significant difference in crashed vehicle rates (See Table 5).

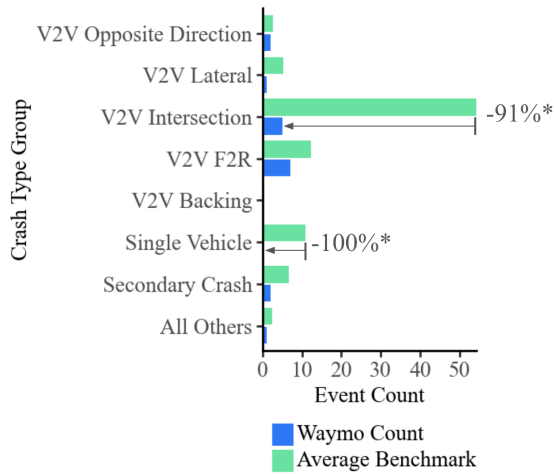

**Figure 4.** Comparison of Observed Waymo and Average Benchmark *Airbag Deployment* Crashes in All Locations over 56.7 Million Miles. Comparisons labeled with an asterisk (*) had a statistically significant difference in crashed vehicle rates (See Table 6).

## DISCUSSION

### Interpretation of Results

The results of this study show for the first time using retrospective crash data that the Waymo RO SAE level 4 ADS has a statistically significant reduction in a *Suspected Serious Injury+* outcome, in addition to crashes resulting in injury of any severity which are dominated by frequent minor injuries. The result of a 85% reduction in *Suspected Serious Injury+* crashes (39% to 99% reduction 95% confidence interval) is an indication of an effect, but is subject to a low number of observations (2 ADS crash) with large confidence intervals. The 2 *Suspected Serious Injury+* crashes involving Waymo vehicles are summarized in the appendix, all of which were of the Secondary Crash crash type. In one of the two crashes, the Waymo vehicle was stationary in traffic when the crash occurred and the other the Waymo vehicle was in a Secondary Crash with a high speed red light runner which was redirected and struck pedestrians on the sidewalk after the crash. As discussed below, future research could develop objective measures of crash contribution that can be applied to both benchmark and ADS crashes. The magnitude of the effect is similar to past simulations studies of the Waymo Driver's performance in reconstructed fatal human crashes (Scanlon et al 2021). The previous simulation study was performed in Chandler, AZ, which is a more suburban location compared to the current operating areas that have the most driving in densely populated areas. This *Suspected Serious Injury+* result is also in line with the multiple complimentary design-based methods used to set requirements and evaluate the Waymo Driver prior to deployment (Webb et al 2020). These methods include simulation-based methods, including the Collision Avoidance Testing (CAT) method (Kusanon et al 2023) that compares the Waymo Driver's performance to a Non-Impaired Eyes On conflict (NIEON) model (Engstrom et al 2024a). Future studies could continue to study the retrospective performance of the *Suspected Serious Injury+* and other high severity outcomes as more ADS mileage is collected that enables further analysis.

When analyzing the crash performance by crash type, the Waymo RO service had statistically significant reductions in Cyclist, Motorcycle, Pedestrian, Secondary Crash, Single Vehicle, V2V Intersection, and V2V Lateral crashes for the *Any-Injury-Reported* outcome and V2V Intersection crashes for the *Airbag Deployment* outcome. Crashes involving VRUs (including Motorcyclists and Pedestrians), Single Vehicle, and V2V Intersection account for a large proportion of the benchmark crash rate for the *Any-Injury-Reported* and *Airbag Deployment* outcome levels. By significantly reducing crashes in these frequent crash modes, the Waymo RO service was able to achieve overall reductions in aggregate crash rates in these two outcomes. Comparisons in all other crash types for these two outcomes were not statistically significant, although generally lower for the SAE level 4 ADS. These non-significant results suggest the need to accumulate more ADS miles in order to determine whether Waymo RO and benchmark crash rates differ. Given the low number of observed *Suspected Serious Injury+* ADS crashes, more miles are needed to draw statistical conclusions about *Suspected Serious Injury+* performance in individual crash types. Taken together, the methods and data examined in this study represent the most comprehensive attempt to account for multiple confounding factors between ADS and benchmark data and features the largest dataset analyzed to-date. Therefore, the results of this study represent the most compelling evidence of a meaningful safety benefit of the Waymo RO SAE level 4 ADS operating in a ride-hailing setting in San Francisco, Los Angeles, and Phoenix.

One important reason for performing analysis by crash type is to isolate crashes with similar contributing factors and thus draw some conclusions about the safety system's performance in those types of crashes. The severity of collisions can differ dramatically between different crash types. Using US crash data, Kusano et al (2023) found that front-to-rear crashes made up 36% of police-reported crashes resulting in property damage or minor injury but only 8% of fatal crashes, while crashes between passenger vehicles and pedestrians or motorcyclists made up 36% and 13% of fatal crashes, respectively, but less than 1% each of minor injury crashes each. Intersection crashes are common in both minor injury (25%) and fatal crashes (27%). Therefore, it is particularly promising that the Waymo RO service had reductions in crash types associated with serious injuries (V2V Intersection, Motorcycle, and Pedestrian crash types). Although there are not yet sufficient miles to analyze the *Suspected Serious Injury+* outcome by crash type, it is promising to see an apparent reduction in the number of *Suspected Serious Injury+* outcome at the aggregate level.

Although there were no statistically significant results suggesting that the Waymo RO service had an elevated crash rate relative to the benchmark in any of the 11 crash modes examined, a supplemental analysis that split the F2R crash type into F2R Striking and F2R Struck rates found the Waymo vehicle had a lower F2R Striking rate for *Airbag Deployment* and *Any-Injury-Reported* at a statistically significant level and a statistically significant increase in F2R Struck crashes at the *Any-Injury-Reported* outcome level in Phoenix (see appendix). As more data becomes available, more statistically significant conclusions will be drawn, and thus it stands to reason how one should interpret an ADS with no change or increase in certain types of crash types but decreases in others relative to some benchmark. This study did not account for crash contribution. Different crash types may have different levels of contribution from parties depending on the maneuver of the party. Another possibility is that different crash types have different proportions of crashes where one or more parties have little to no opportunity to contribute to the outcome of the crash. For example, in Scanlon et al (2021) which simulated the Waymo Driver placed in reconstructed fatal crashes involving human drivers, in 8% of responder role simulations there was little to no opportunity for the Waymo Driver to avoid the collision (i.e., the vehicle was stopped at a traffic light when struck from behind at high rates of speed). This type of collision where the Waymo ADS vehicle was stationary in traffic with little to no ability to prevent the collisions accounted for 1 out of the 2 *Suspected Serious Injury+* crashes during the study period (30270-9724).

Regardless of crash contribution, past road safety systems, such as automated red light enforcement cameras and front crash prevention systems, have shown the ability to reduce a certain type of collision while slightly increasing the rate of a different type of collision. Because road safety, and the harm that result from traffic crashes, is often seen as a public health issue, systems are judged on their aggregate, overall contribution to safety, and some risk redistribution is tolerated if the aggregate safety benefit is deemed acceptable. For example, red light enforcement cameras (infrastructure that can automatically detect and ticket drivers who run red traffic lights) have been found to reduce fatal red light running and all fatal crashes at intersections after installation, but increase rear-end collisions to a lesser degree in some situations (McGee and Eccles 2003, Hu and Cicchino 2017). The main safety intervention (to disincentivize red light running) was effective in reducing intersection crashes, but may have also caused some drivers to suddenly come to a stop to avoid the potential of running a red light thus increasing the risk of being struck in the rear. Because red light running crashes, which feature vehicles often traveling at high speeds and side impacts which are more injurious than front-to-rear impacts, were reduced to a large degree, the potential increase of rear-end crashes was accepted because of the overall benefit of this safety intervention. Similarly, a large retrospective study using insurance claims data for front crash prevention systems, including FCW and AEB, found a combination of FCW and AEB reduced crashes by 50% and injury crashes by 56%, but increased rear-end struck crashes by 20% (Cicchino 2017). Overall, these front crash prevention systems were estimated to have a potential to prevent 1 million crashes per year and 400,000 injuries (Cicchino 2017), which is one of the reasons 20 automakers voluntarily committed to making AEB standard equipment on all new vehicles in the US by 2022. These are just two examples of numerous other road safety innovations that have reduced overall harm but had some measurable residual risks. Therefore, ADS should be considered in a similar way as a replacement for human driving where the evidence suggests there are large aggregate safety benefits. Potential increases over a benchmark in sub-groups of crashes should be considered in the context of the overall safety benefit of the system as well as the role of crash contribution, which is not considered in this study.

Similarly, the current study found the Waymo RO service had large reductions in V2V Intersection, Single Vehicle, and VRU crashes. In all locations and compared to a driver with the average benchmark rate driving the same 56.7 million miles as the Waymo Driver, the Waymo Driver experienced 86 fewer V2V Intersection, 45 fewer VRU, 14 fewer Single Vehicle, 12 fewer V2V Lateral, 10 fewer V2V F2R, and 8 fewer Secondary Crash *Any-Injury-Reported* Crashes. Compared to an average human driver, the Waymo RO service experienced 50 fewer

V2V Intersection, 11 fewer Single Vehicle, and 5 fewer F2R *Airbag Deployment* Crashes. When breaking down F2R crashes by F2R Striking and F2R Struck, the Waymo RO service had 17 fewer F2R Striking and 8 more F2R Struck *Any-Injury-Reported and* crashes and 6 fewer F2R Striking and 1 more F2R Stuck *Airbag Deployment* crashes compared to an average human. Like past safety systems, the magnitude reductions of the Waymo RO service in most crash groups, including those that most often result in the most serious injuries, were far larger than the increases in F2R Struck crashes.

See the appendix for additional results comparisons to prior studies.

**Difficulties in Examining ADS Crash Contribution**

The research questions addressed in the current study are related to the Waymo ADS overall crash rate compared to human benchmarks, regardless of contribution. This overall crash rate view compliments past studies that use 3rd party liability claims rate as a surrogate for ADS crash contribution and have found the Waymo RO service has a large reduction in 3rd party property damage and personal injury liability claims (Di Lillo et al 2024a, 2024b). Even 3rd party liability claims have limitations in that frequency and/or payment amounts associated with insurance claims may not always be due to responsibility in the collision. Additionally, companies that operate fleets of vehicles like the current ride-hailing ADS deployments may have different insurance risk profiles as private insurance that is used by many human drivers. Although insurance claims data likely is a good proxy to investigate ADS contribution in crashes, there is a research need to develop more objective assessments of crash contribution. One possible way to develop such objective crash contribution assessments is through the use of reference behavior models, sometimes called driver models. The reference models used for crash contribution assessments should reflect proper Drivership, including behavior that matches normative expectations of good driving (Fraade-Blanar et al 2025).

**Limitations**

This study had several limitations. Although the study takes steps to align the benchmark and ADS data using dimensions available in both human and ADS data sources (such as geographic location, road type, vehicle type, and spatial driving density), there is an endless list of possible factors to potentially account for and of adjustment methodologies to refine. One dimension discussed in the dynamic benchmark adjustment done by Chen et al (2024), but not implemented in current study, is an adjustment for time of day. Chen et al (2024) found that the Waymo ADS fleet, through the first 21.9 million RO miles, that the ADS fleet in San Francisco drove slightly more in the evening and overnight (6:30PM - 3AM, 41% vs 24%) and early morning (3AM - 6AM, 6% vs 3%) and less during the daytime (9AM - 3:30PM, 16% vs 40%). In total, however, the time-of-day adjustment resulted in a benchmark that was higher than the baseline benchmark by 1.05 times [1.03, 1.06] for *Any-Injury-Reported* outcome and 1.16 times [1.13, 1.18] for the *Airbag Deployment* outcome in San Francisco (Chen et al 2024). This suggests that the benchmark in San Francisco used in this study is conservative, in that it likely underestimates elevated crash risk with driving more at higher risk times of day. This time-of-day adjustment was only possible in San Francisco based on extrapolating a single traffic study that had human VMT by time-of-day. No such data could be found in other cities, and this single study is likely less robust than the spatial VMT data used by Chen et al (2024) to perform the spatial dynamic adjustment used in the current study. This study used an underreporting adjustment for the benchmark *Any-Injury-Reported* outcome, which accounts for the 33% of injury crashes that were estimated to not be reported to police by Blincoe et al (2023). The *Airbag Deployment* and *Suspected Serious Injury+* benchmarks, however, did not have an underreporting adjustment applied even though there is likely non-zero underreporting in human crash data. This lack of underreporting adjustment for these outcomes likely makes the comparisons in these levels conservative, because there is assumed to be little underreporting in the ADS crash data due to automated collision reporting using vehicle sensors and operational procedures (Kusano et al 2024).

Aligning the benchmark and Waymo crash and mileage data also comes with its implementation challenges. On the Waymo data side, the authors have detailed information about the mileage and crashes that have been made available. On the human driving side, there is uncertainty and potential bias introduced due to the nature of the mileage and crash data. A variety of geographic-specific variable and value pairing are needed to do this study's classification routines for vehicle type, road type, and crash type (see the appendix). The underlying raw data being relied upon (e.g., police reports) have limited specificity and are also subject to input error. The mileage estimates are also derived from a variety of geographic-specific traffic sampling methodologies, and, in the absence of some ground truth data to validate, it is not clear how much uncertainty or bias should be attributed to these estimates. In the crash type groups used in this paper, there was an "All Others" category to capture both unknown crash types and crash types that have a configuration that is not described by the other 10 crash type groups. Only the human data had "unknown" crash type, as sufficient details to determine crash type were present for all SGO crashes examined. Examples of crashes included in the "All Other" category include "vehicle-to-vehicle dooring crashes" (where the open door of a vehicle is struck by another road user) and other unique collision circumstances that did not fit into the existing 10 categories.

Scanlon et al (2024) examined the benchmark crash rates over time and found that after a disruption during the COVID-19 pandemic in 2020, the 2021 and 2022 crash rates were relatively stable. As most of the Waymo RO driving was performed in 2023 and beyond (98% of the miles), future work could examine the effect of changes in the human benchmark with time.

Similar to the analysis in Kusano et al (2024), the Waymo ADS vehicle was not always occupied while driving in RO configuration (e.g., traveling between dropping off and picking up passengers). The research question of the current study investigates the effect of the Waymo RO service on the current status quo of human driving. As Waymo is operating a ride-hailing service at relatively small scales relative to the overall human driving population, it is not unreasonable to assume that much of the VMT driven by the Waymo RO service would have been serviced by human ride-hailing services. In human ride-hailing, there is a human driver driving the vehicle between dropping off and picking up passengers with the potential to be injured if a crash occurs. The Waymo service thus reduces the risk of injury by removing additional people from potentially hazardous crashes. Additionally, the analysis includes any injury to any participant involved in a crash, including other human-operated vehicles involved in the crash. Even if the Waymo vehicle was completely unoccupied at all times, there could still be the potential for human injuries in the event of crashes, especially given that the Waymo mileage is being accumulated in dense urban areas where vulnerable road user crashes are more common. The *Airbag Deployment* outcome was intentionally included as an independent, complementary outcome that is not sensitive to the occupancy status of the Waymo vehicle. The Waymo vehicles comply with all relevant Federal Motor Vehicle Safety Standards (FMVSSs), and thus many of the airbag systems, including the driver front and side airbags, will fire regardless of occupancy. Future research could address the potential impacts of VMT shifts caused by automated driving and shifts in vehicle occupancy and/or vehicle seating position that may affect observed outcomes, but it should be noted that this removes the inherent benefits achieved from a lack of an occupant to be injured in between ride-hailing pickups. Another alternative to examining outcomes is to use injury risk functions that are functions of crash inputs (like delta-V) and are insensitive to occupancy or reported outcomes.

Some human crash databases define injuries using the Abbreviated Injury Scale (AIS) (AAAM 2016). The AIS is an anatomical injury scoring scheme that is coded by certified professionals based on medical records. The ADS crashes reported in the NHTSA SGO and used in this study do not have AIS codes available. Having available AIS, or other more detailed injury data than police-reported maximum severity, would make the ADS SGO more comparable to some human crash data sources that also have AIS injury data. This AIS coding, however, would require additional resources dedicated for crash investigation and a framework to address privacy concerns. Typically, human crash data with AIS codes in the US are overseen by NHTSA and information is collected by independent crash investigation firms, and not self-reported by manufacturers like the NHTSA SGO.

If multiple comparisons are made across many different dimensions, the probability of detecting false positive significant results increases. This study performs a number of comparisons across different outcome levels (3), crash modes (11), and locations (4). The approach to examine multiple outcome levels, which each have unique reporting challenges, was an attempt to draw broad conclusions about the performance of ADS relative to the current human driving population. The crash type comparison in the study was performed to gain insight into which crash modes contribute to the observed aggregate safety benefits. However, due to multiple comparisons, care should be taken when considering the statistical significance of individual comparisons as no multiple comparison corrections have been made.

## CONCLUSIONS

An increase in ADS mileage on public roads in recent years enables additional retrospective safety assessments to be performed. This study compared crash rates of the Waymo RO ride-hailing service to aligned benchmarks in Phoenix, San Francisco, Los Angeles, and Austin for the outcomes of *Any-Injury-Reported*, *Airbag Deployment*, and *Suspected Serious Injury+*. Compared to past ADS safety impact studies, the 56.7 million RO miles and the ADS relative performance to the benchmark during the study period through January 2025 allowed for the first time a statistically relevant comparison to a *Suspected Serious Injury+* outcome at the aggregate (all crash) level and comparisons disaggregated into 11 crash types for the *Any-Injury-Reported* and *Airbag Deployment* outcomes. At the aggregate crash level, the study found statistically significant reductions in *Any-Injury-Reported* and *Airbag Deployment* outcomes when considered in all locations combined (79% CI: [71%, 85%] and 81% CI:[69%, 90%] reduction, respectively) and in San Francisco and Phoenix individually. The ADS reductions in these two outcome groups was of a similar magnitude than an earlier study of the first 7.1 million Waymo RO miles (Kusano et al 2024). The current study found the Waymo RO service had a statistically significant reduction in *Suspected Serious Injury+* outcome crashes compared to the benchmark when considering all locations combined (85% CI: [39%, 99%] reduction), ~~and~~ in Phoenix (100% CI: [3%, 100%]), and in San Francsico (76% CI: [3%, 98%]). Comparisons in Los Angeles were not statistically significant. Both (2) *Suspected Serious Injury+* crashes involving a Waymo vehicle during the study period were Secondary Crashes, meaning the Waymo was not involved in the first event in the crash sequence. Compared to the benchmark crash rate representing the current driving fleet, the Waymo Driver experienced 181 fewer *Any-Injury-Reported*, 78 fewer *Airbag Deployment*, and 11 fewer *Suspected Serious Injury+* crashes during the study period.

The aggregate crash reductions by the Waymo ADS in the *Any-Injury-Reported* outcome group were driven by statistically significant reductions in Cyclist (82%), Motorcycle (82%), Pedestrian (92%), Secondary Crash (66%), Single Vehicle (~~100%~~ 93%), V2V Intersection (96%), and V2V Lateral (74%) crash types when considering all locations. The reduction in Single Vehicle (100% reduction) and V2V Intersection (91% reduction) *Airbag Deployment* crashes was also statistically significant in all locations combined. All other crash rate comparisons disaggregated by crash type found the Waymo RO crash rate was not statistically different from the benchmark rate in all locations combined. The *Airbag Deployment* V2V Intersection comparison was statistically significant and lower in Phoenix, San Francisco, and Los Angeles when examined separately for *Any-Injury-Reported* and *Airbag Deployment* crashes.

The results of this study suggest increasing confidence that a level 4 ADS reduces *Any-Injury-Reported* and *Airbag Deployment* outcome crashes, primarily by reducing V2V Intersection and Single Vehicle crashes for both outcomes and VRU (i.e., cyclist, motorcyclist, and pedestrian), Secondary, and Lateral crashes for the *Any-Injury-Reported* outcome. The crash groups with significant reductions represent some of the most frequent crash modes in the environment the Waymo ADS currently operates. The results of this study strongly suggest a safety benefit of the Waymo RO service, an SAE level 4 ADS, operating as a ride-hailing vehicle in the *Any-Injury-Reported* and *Airbag Deployment* outcome levels, and suggest a benefit for *Suspected Serious Injury+* crashes. The methodology of this study crafted the benchmarks through subselection and adjustments that attempted to account for many known factors that affect crash risk. Because conservative assumptions were made where appropriate and sensitivity analysis was performed, the remaining uncertainty in the benchmarks would likely not change the conclusions of the study. Future research should continue refining the alignment between benchmark and ADS crash and mileage data sources and continue to monitor serious and fatal severity injury outcomes, like the *Suspected Serious Injury+* outcome, which have traditionally been the primary focus of road safety efforts.

**DISCLOSURE STATEMENT**

All authors are employed by Waymo LLC.

**DATA AVAILABILITY STATEMENT**

Public data sources used in this analysis are cited and available from state agencies and NHTSA. Waymo mileage data is self-reported and available for download at https://www.waymo.com/safety/impact. Data files with full study results and listing of NHTSA SGO cases used in this analysis are provided as supplemental online downloads.

**APPENDIX**

**ADS and Benchmark Crash Rate Definitions**

All ADS operators in the US, including Waymo, are required to report collisions as part of the NHTSA SGO (NHTSA 2023). The SGO requires reporting of "any physical impact between a vehicle and another road user (vehicle, pedestrian, cyclist, etc.) or property that results or allegedly results in any property damage, injury, or fatality." This reporting extends to crashes where the ADS vehicle was impacted or in cases where the ADS "contributes or is alleged to contribute (by steering, braking, acceleration, or other operational performance) to another vehicle's physical impact with another road user or property involved in that crash." Benchmark data was not universally available for the subset of crashes where some third party non-contacted vehicle may have contributed to some crash outcome, so this study excluded crash events without direct contact with the Waymo. This limited analysis scope and opportunity for future research is discussed later in the document.

At this time, there are no universal state or federal requirements to report ADS operation driving miles. California requires various mileage reporting overseen by the Department of Motor Vehicles (DMV) and Public Utilities Commission (PUC). There are no VMT reporting requirements outside of California. Because of this data limitation, Waymo voluntarily releases driving mileage by deployment geographic region in a downloadable dataset (Waymo 2025). Because other ADS operator miles are currently unavailable, this study restricted its analysis to only examine Waymo's safety performance. Furthermore, because this study only examined Waymo RO operations, only driving miles and crashes without an autonomous specialist sitting in the driver's seat overseeing the ADS were analyzed.

**Human Benchmark Data**

**Crash Data Sources:** Publicly-available, police-reported crash data from the states of Arizona, Texas, and California were used to generate geographic region specific benchmark crash count estimates. The Arizona Department of Transportation (ADOT) and Texas Department of Transportation (TxDOT) both annually compile a census of all police-reported crashes within their state. The California Highway Patrol (CHP) compiles a subset of the total police-reported crashes within the Statewide Integrated Traffic Records System (SWITRS). In accordance with California Vehicle Code §20008, SWITRS is intentionally designed to capture all the crashes that involved an injured person, where crashes involving property damage only (PDO) are not universally reported as "some agencies report only partial numbers of their PDO crashes, or none at all" (CHP 2021).

**Mileage Data Sources:** Waymo mileage totals and distributions are established through continuous tracking of sensors (e.g., vehicle odometer readings on the rear wheel). Publicly-available mileage estimates were obtained for Arizona, Texas, and California that are listed in Table 1. Each of these states annually report the total miles driven across the counties of the state by subsets of road types. All of these mileage totals are geographic-specific estimates based on established traffic counting sampling methodologies on specific roadways for limited periods within the respective jurisdictions. The extrapolation to all roadways is done using geographic-specific modeling techniques. The Federal Highway Administration (FHWA) highway statistics data VM-4 tables for urban roadways were used to adjust the state mileage estimates by vehicle type (FHWA 2023a). Additionally, for the described spatial adjustments, the FHWA-compiled Highway Performance Monitoring System (HPMS) was used to get road level mileage estimates (FHWA 2016). This database contains Annual Average Daily Traffic (AADT) estimates for all roadways in the analyzed geographic regions with the exception of local roads and rural minor collectors.

**Subselection of In-Transit Passenger Vehicles on Surface Streets:** This study used the same classification routine as was used in Scanlon et al. (2024a) for identifying crashes and mileage associated with surface street, in-transit passenger vehicles for both California and Arizona data. Additionally, these methods have been extended to cover Texas. Table A1 provides a full accounting of the data data fields relied upon for doing this subselection.

Table A1. VariablesUsed to Subset Public Crash and Mileage Data for in-Transport, passenger vehicles on Surface Streets.

| State | Data Source | Variables used | | |
|---|---|---|---|---|
| | | Surface Street | In-Transport | Passenger Vehicles |
| California, Arizona, Texas | FHWA Highway Statistics Series (VM-4 Table) | "other arterials", "other" | All Mileage | "passenger cars", "light trucks" |
| | FHWA Highway Performance Monitoring System | f_system | All Mileage | aadt, aadt_singl, aadt_combi |
| California | CHP SWITRS | chp_beat_type | Move_pre_acc, Party_type | party_type, Stwd_vehicle_type, Chp_veh_type_towing |
| | Public Road Data | Excludes interstates, state routes, and US highways | All Mileage | (applies FHWA VM-4 correction) |
| Arizona | ADOT Crash Data | GeocodeOnRoad, PostedSpeed | UnitAction | BodyStyle, UnitType |
| | ADOT Certified Public Miles | Excludes "Interstates" and "Other Arterials - Other Freeways and Expressways" | All Mileage | (applies FHWA VM-4 correction) |
| Texas | TXDOT CRIS | Rpt_Street_Sfx, Rpt_Rdwy_Sys_ID, Rpt_Street_Name, Rpt_Street_Desc, | Veh_Parked_Fl | VIN*, Veh_Body_Styl_ID, Cmv_Veh_Type_ID, Unit_Desc_ID, Cmv_GVWR, |

| | | Func_Sys_ID, Rpt_Hwy_Num, Crash_Speed_Limit | | Cmv_Fiveton_Fl |
|---|---|---|---|---|
| | **TXDOT Roadway Inventory Annual Reports** | Excludes "Interstates" and "Other Freeway-Expressway" | All Mileage | (applies FHWA VM-4 correction) |

\* VINs were decoded using NHTSA's publicly-available VIN decoder API (https://vpic.nhtsa.dot.gov/api/vehicles/DecodeVinValues/).

Some vehicle types were not specified with enough specificity to determine whether the actor was a passenger vehicle or some other actor type (e.g., a heavy vehicle). For these uncertain vehicles, the vehicle type was imputed by applying a weighting factor representative of how likely that actor was to be a passenger vehicle. This weight was calculated using the proportion of the known vehicles that were a passenger vehicle for each specific geographic region. The weighting factors applied to the unspecified vehicles were 0.93, 0.89, 0.89, and 0.94 for Los Angeles, San Francisco, Maricopa County, and Austin, respectively.

**Subselection By Crash Type:** This study presented a new methodology to determine crash type from available data. These crash types were not directly indicated in the data itself, but, rather, were inferred using the available data. The first step was assigning each actor in the collision a body type. The second step determined whether the actor being counted was a secondary collision partner not involved in the primary collision event. Those actors received their own crash type classification. The third step coded the primary collision partners according to the type of actors involved and the indicated crash configuration. The relied upon variables for each state crash database is provided in Table A2.

Table A2. Variables used to subset identify actor type and crash type

| | | **Variables used** | |
|---|---|---|---|
| **State** | **Data Source** | **Actor Type** | **Crash Type (in addition to variables in actor type)** |
| **California** | **CHP SWITRS** | Party_type, Stwd_vehicle_type, Chp_veh_type_towing | party_number, pcf_violation, move_pre_acc, dir_of_travel, type_of_collision |
| **Arizona** | **ADOT Crash Data** | UnitAction, BodyStyle, UnitType | party_count, EventSequence1, CollisionManner, JunctionRelation, UnitAction |
| **Texas** | **TXDOT CRIS** | Veh_Parked_Fl, | FHE_Collsn_ID, |

| | | VIN*, Veh_Body_Styl_ID, Cmv_Veh_Type_ID, Unit_Desc_ID, Cmv_GVWR, Cmv_Fiveton_Fl | Contrib_Factr_1_ID, Contrib_Factr_2_ID, Contrib_Factr_3_ID |
|---|---|---|---|

**Subselection of Outcome Levels:** This study applied the same outcome level coding scheme that was applied in Scanlon et al. (2024a). The current study does extend those previously described methods to Texas. Table A3 provides the variables relied upon for subselection outcome levels.

Table A3. Variables used to identify outcome level.

| State | Data Source | Variables used | |
|---|---|---|---|
| | | **Any-Injury Reported & Suspected Serious Injury+** | **Airbag Deployment** |
| **California** | **CHP SWITRS** | collision_severity | Party_safety_equip_1, party_safety_equip_2, Victim_safety_equip_1, victim_safety_equip_2 |
| **Arizona** | **ADOT Crash Data** | InjuryStatus | Airbag |
| **Texas** | **TXDOT CRIS** | Crash_Sev_ID | Prsn_Airbag_ID |

The *any-injury-reported* crash level also received an underreporting correction using NHTSA's most recent estimate on underreporting of injury crashes in the United States (Blincoe et al., 2023). Blincoe et al. (2023) estimated that 60% of PDO crashes and 32% of non-fatal injury crashes were not reported to police. Their estimates were produced through multiple steps of computation leveraging telephone surveyed data, insurance records, police reports, and other sources. In practice, this technique applies a 1.47 weighting factor to the non-fatal injury crashes.

This underreporting estimate is not specific to the geographic area being evaluated, i.e., the extent of underreporting is being estimated from this national data. It is noteworthy that a study by the San Francisco Department of Public Health (SFDPH) and the San Francisco Municipal Transportation Agency (SFMTA) estimated that "severe" injuries (defined as an injury severity score (ISS) of 15 or greater) were not reported by police for 39% of vehicle occupants, 28% of pedestrians, and 33% of cyclist injuries. So, in San Francisco, we expect the Blincoe correction applied to be conservative.

Another dimension examined in this study was whether the ADS vehicle was the initiator or responder in the crashes (Kusano et al., 2023). In a crash or conflict that involves two parties, there is an initiator, which is the actor that performs the initial unexpected or surprising behavior that leads to the conflict occurring. The other party is the responder, who needs to respond to the surprising actions of the initiator. For example, a vehicle that runs a red light and crashes into another vehicle going in the perpendicular direction would be the initiator role. Initiator and responder role designation was annotated for crashes based on manual review of the ADS sensor and video data. Examining the initiator and responder role of the ADS in individual crash modes can also be useful in understanding potential differences to the human crash population. Although there are limited human crash benchmarks that have initiator and responder role annotated, in general in a crash between two humans there should be one initiator and one responder, leading to an approximate 50% initiator and responder split.

**Waymo SGO Data**

*Airbag Deployment* crashes were those where an airbag deployed in any involved vehicle. Even in ADS crashes without a passenger in the vehicle, the frontal (driver-side) and side airbags will deploy, which enables a comparative benchmark to the human-driven population. In the NHTSA SGO data, an *Airbag Deployment* crash was identified by either the field "SV Any Air Bags Deployed?" or "CP Any Air Bags Deployed?" was reported as "Yes". Additionally, video data was reviewed from all crashes to determine if an airbag was deployed in any vehicle involved in the crash sequence, regardless of whether that vehicle was the primary crash partner of the Waymo vehicle as reported in the SGO. *Any-Injury-Reported* crashes were those where the SGO field "Highest Injury Severity Alleged" was reported as "Minor", "Moderate", "Serious", or "Fatality". Additionally, any case with the field "Highest Injury Severity Alleged" with a value of "Unknown" and mention of an injury of unknown severity in the SGO case narrative was also included in the *Any-Injury-Reported* outcome. This *Any-Injury-Reported* definition is the same used in Kusano et al. (2024). The Suspected Serious Injury+ outcome was a subset of the Any-Injury-Reported with the field "Highest Injury Severity Alleged" having a value of either "Serious" or "Fatality". This definition aligns with the *Suspected Serious Injury+* benchmark from Scanlon et al. (2024), where the police-reported KABCO scale values of K and A (killed and incapacitating injury, respectively) were used.

Identifying Waymo in-transport mileage and crashes on surface streets required some data subselection. Both in-transport status and surface streets are made publicly available on Waymo's public crash dashboard for public analysis (Waymo 2025). A not in-transport status meets two conditions. First, the vehicle must be in the "park" gear. Second, the vehicle must be within a designated parking space. If parked against a curb, the ADS must have been within 18 inches of the curb edge, which was measured from on-board lidar data. This in-transport definition is the same that was used in Kusano et al (2024).

**Additional Results**

Select additional results are shown in this appendix section. The supplemental downloads of this paper also includes the file "CSV3 - Collision Counts and Comparison to Benchmarks 202009-202501-2022benchmark.csv", which contains even more comparisons of the Waymo RO crash rate to human benchmarks. The data used in the dynamic benchmark calculation is provided as a supplemental download with file name "CSV4 - Miles and Benchmark Crashes for Dynamic Benchmark 202009-202501-2022benchmark.csv". The data dictionary for these supplemental downloads can be found here.

Table A4. Comparison of Waymo RO and Human Benchmark Crashed Vehicle Rates for *Any-Injury-Reported*, *Airbag Deployment*, and *Suspected Serious Injury+* Crashes in Los Angeles.

| Outcome | Human IPMM | ADS IPMM | Expected ADS Count Different to Benchmark | ADS to Human Percent Difference | 95% Confidence Interval | |
|---|---|---|---|---|---|---|
| *Any-Injury-Reported* | 2.37 | 1.24 | -7.3 | -48% | -80% | 12% |
| *Airbag Deployment* | 1.18 | 0.31 | -5.6 | -74% | -98% | 7% |
| *Suspected Serious Injury+* | 0.14 | 0.00 | -0.9 | -100% | -100% | 300% |

Table A5. Number of Crashes by Outcome and Crash Type in All Locations (through January 2025, 56.7M RO Miles).

| Crash Type | Any-Injury-Reported | Airbag Deployment | Suspected Serious Injury+ |
|---|---|---|---|
| Cyclist | 3 | 0 | 0 |
| Motorcycle | 2 | 0 | 0 |
| Pedestrian | 2 | 0 | 0 |
| Secondary Crash | 4 | 2 | 2 |
| Single Vehicle | 1 | 0 | 0 |
| V2V Backing | 0 | 0 | 0 |
| V2V F2R | 25 | 7 | 0 |
| V2V Opposite Direction | 2 | 2 | 0 |
| V2V Intersection | 4 | 5 | 0 |
| V2V Lateral | 4 | 1 | 0 |
| All Other | 1 | 1 | 0 |

Table A6. Comparison of ADS and Human Benchmark (with Dynamic Benchmark Adjustment) Crashed Vehicle Rates in Phoenix and San Francisco by Crash Type in *Any-Injury-Reported* Crashes (Blincoe-Adjusted Benchmark) (through January 2025, 56.7M RO Miles).

| Location | Crash Type | Human IPMM | ADS IPMM | Expected ADS Count Different to Benchmark | ADS to Human Percent Difference | 95% Confidence Interval | |
|---|---|---|---|---|---|---|---|
| Phoenix | Cyclist | 0.04 | 0.00 | -1.3 | -100% | -100% | 185% |
| | Motorcycle | 0.05 | 0.06 | 0.4 | 25% | -90% | 409% |
| | Pedestrian | 0.07 | 0.00 | -2.3 | -100% | -100% | 60% |
| | Secondary Crash | 0.17 | 0.03 | -4.2 | -81% | -100% | 23% |
| | Single Vehicle | 0.12 | 0.00 | -3.6 | -100% | -100% | 2% |
| | V2V Backing | 0.00 | 0.00 | -0.1 | -100% | -100% | 2685% |
| | V2V F2R | 0.44 | 0.48 | 1.2 | 9% | -44% | 90% |
| | V2V Opposite Direction | 0.06 | 0.03 | -0.9 | -48% | -99% | 235% |
| | V2V Intersection | 0.97 | 0.10 | -27.1 | **-90%*** | **-98%** | **-68%** |
| | V2V Lateral | 0.13 | 0.06 | -2.0 | -49% | -96% | 106% |
| | Other | 0.04 | 0.00 | -1.2 | -100% | -100% | 204% |
| San Francisco | Cyclist | 0.80 | 0.11 | -12.5 | **-86%*** | **-99%** | **-44%** |
| | Motorcycle | 0.52 | 0.00 | -9.4 | **-100%*** | **-100%** | **-61%** |
| | Pedestrian | 1.12 | 0.06 | -19.5 | **-95%*** | **-100%** | **-69%** |
| | Secondary Crash | 0.31 | 0.16 | -2.7 | -47% | -92% | 74% |
| | Single Vehicle | 0.56 | 0.00 | -10.1 | **-100%*** | **-100%** | **-63%** |
| | V2V Backing | 0.10 | 0.00 | -1.7 | -100% | -100% | 118% |
| | V2V F2R | 0.97 | 0.44 | -9.6 | **-55%*** | **-83%** | **-3%** |
| | V2V Opposite Direction | 0.08 | 0.00 | -1.4 | -100% | -100% | 179% |
| | V2V Intersection | 2.84 | 0.00 | -51.8 | **-100%*** | **-100%** | **-93%** |
| | V2V Lateral | 0.56 | 0.11 | -8.3 | **-81%*** | **-98%** | **-20%** |
| | Other | 0.18 | 0.00 | -3.3 | -100% | -100% | 12% |
| Los Angeles | Cyclist | 0.10 | 0.16 | 0.4 | 61% | -98% | 927% |

| Location | Crash Type | Human IPMM | ADS IPMM | Expected ADS Count Different to Benchmark | ADS to Human Percent Difference | 95% Confidence Interval | |
|---|---|---|---|---|---|---|---|
| | Motorcycle | 0.06 | 0.00 | -0.4 | -100% | -100% | 850% |
| | Pedestrian | 0.15 | 0.16 | 0.0 | 5% | -99% | 569% |
| | Secondary Crash | 0.16 | 0.00 | -1.0 | -100% | -100% | 251% |
| | Single Vehicle | 0.19 | 0.16 | -0.2 | -17% | -99% | 427% |
| | V2V Backing | 0.02 | 0.00 | -0.1 | -100% | -100% | 3336% |
| | V2V F2R | 0.40 | 0.31 | -0.5 | -21% | -93% | 219% |
| | V2V Opposite Direction | 0.04 | 0.16 | 0.8 | 311% | -95% | 2528% |
| | V2V Intersection | 1.04 | 0.16 | -5.7 | **-85%*** | **-100%** | **-5%** |
| | V2V Lateral | 0.17 | 0.00 | -1.1 | -100% | -100% | 233% |
| | Other | 0.05 | 0.16 | 0.7 | 208% | -96% | 1870% |

Table A7. Comparison of ADS and Human Benchmark (with Dynamic Benchmark Adjustment) Crashed Vehicle Rates in Phoenix and San Francisco by Crash Type in *Airbag Deployment* Crashes (through January 2025, 56.7M RO Miles).

| Location | Crash Type | Human IPMM | ADS IPMM | Expected ADS Count Different to Benchmark | ADS to Human Percent Difference | 95% Confidence Interval | |
|---|---|---|---|---|---|---|---|
| Phoenix | Secondary Crash | 0.12 | 0.00 | -3.8 | **-100%*** | **-100%** | **-2%** |
| | Single Vehicle | 0.10 | 0.00 | -3.2 | -100% | -100% | 14% |
| | V2V Backing | 0.00 | 0.00 | 0.0 | -100% | -100% | 11612% |
| | V2V F2R | 0.24 | 0.16 | -2.4 | -32% | -82% | 73% |
| | V2V Opposite Direction | 0.05 | 0.03 | -0.6 | -38% | -99% | 295% |
| | V2V Intersection | 0.81 | 0.06 | -23.2 | **-92%** | **-99%** | **-68%** |
| | V2V Lateral | 0.07 | 0.00 | -2.0 | -100% | -100% | 81% |
| | Other | 0.03 | 0.00 | -0.9 | -100% | -100% | 319% |
| San Francisco | Secondary Crash | 0.13 | 0.06 | -1.3 | -56% | -99% | 187% |
| | Single Vehicle | 0.35 | 0.00 | -6.4 | **-100%*** | **-100%** | **-42%** |
| | V2V Backing | 0.02 | 0.00 | -0.3 | -100% | -100% | 1480% |

| Location | Crash Type | Human IPMM | ADS IPMM | Expected ADS Count Different to Benchmark | ADS to Human Percent Difference | 95% Confidence Interval | |
|---|---|---|---|---|---|---|---|
| | V2V F2R | 0.20 | 0.06 | -2.7 | -73% | -100% | 75% |
| | V2V Opposite Direction | 0.04 | 0.06 | 0.2 | 27% | -98% | 768% |
| | V2V Intersection | 1.31 | 0.11 | -21.9 | **-92%*** | **-99%** | **-66%** |
| | V2V Lateral | 0.15 | 0.06 | -1.7 | -63% | -100% | 138% |
| | Other | 0.07 | 0.06 | -0.3 | -21% | -99% | 429% |
| Los Angeles | Secondary Crash | 0.09 | 0.16 | 0.4 | 76% | -98% | 1023% |
| | Single Vehicle | 0.18 | 0.00 | -1.1 | -100% | -100% | 226% |
| | V2V Backing | 0.01 | 0.00 | 0.0 | -100% | -100% | 11329% |
| | V2V F2R | 0.13 | 0.16 | 0.2 | 19% | -99% | 661% |
| | V2V Opposite Direction | 0.02 | 0.00 | -0.1 | -100% | -100% | 2402% |
| | V2V Intersection | 0.63 | 0.00 | -4.1 | **-100%*** | **-100%** | **-10%** |
| | V2V Lateral | 0.07 | 0.00 | -0.5 | -100% | -100% | 699% |
| | Other | 0.03 | 0.00 | -0.2 | -100% | -100% | 1910% |

**Underreporting and Dynamic Benchmark Sensitivity Study**

To determine the sensitivity of the findings of this study to underreporting correction used in the *Any-Injury-Reported* human benchmark and to the dynamic benchmark adjustment, we conducted a sensitivity analysis. The underreporting correction and dynamic benchmark were implemented to better align the human benchmark and Waymo RO crash data based on factors known to affect crash rates. Human data is known to have underreporting, even for injury crashes. The driving environment within a geography can also affect crash rates, and the Waymo RO service travels in higher-density (and higher crash rate) areas more frequently than the overall human driving population. The purpose of these sensitivity analyses is to bound the size of the effect of these benchmark adjustments and determine if the overall conclusions of the study would be changed if these adjustments were not made. Due to the large number of comparisons involved, individual data are not tabulated in this appendix section. All results discussed can be found in the supplemental download file "CSV3 - Collision Counts and Comparison to Benchmarks 202009-202501-2022benchmark.csv" described in the previous section.

Similar to the sensitivity analysis performed by Kusano et al (2024) on the first 7.1 million Waymo RO miles, the conclusions for the reduction in all crashes in the *Any-Injury-Reported* outcome in the current study were not changed due to the 33% underreporting assumption used to adjust the benchmark. Using an observed

*Any-Injury-Reported* benchmark with dynamic benchmark adjustment also found a significant reduction in *Any-Injury-Reported* crashes in Phoenix, San Francisco, and in all locations combined. This is the same result as when using the underreporting adjusted benchmark. Furthermore, the same statistically significant reductions are observed if the Waymo RO crash rate is compared to the observed (no underreporting correction) benchmark without a dynamic benchmark adjustment. Because *Any-Injury-Reported* crashes happen at a high rate relative to the miles driven, even when these underreporting and dynamic adjustments are not included (which bias the data toward the conservative), the Waymo RO service still exhibits a statistically significant reduction in *Any-Injury-Reported* crashes overall.

Similarly, the dynamic benchmark adjustment did not change the conclusions of the analysis for all crashes in the *Any-Injury-Reported* and *Airbag Deployment* outcomes. When comparing the Waymo RO crash rate to *Any-Injury-Reported* and *Airbag Deployment* benchmarks without a dynamic benchmark adjustment, the all crash rate reductions were also significant in Phoenix, San Francisco, and all locations combined. The dynamic benchmark adjustment for the *Suspected Serious Injury+* outcome did affect the statistical significance of the comparison of Waymo and the benchmark. Without the dynamic benchmark adjustment, the reductions in *Suspected Serious Injury+* crashes was not statistically significant in Phoenix or San Francisco but was statistically significant in all locations combined, whereas it was statistically significant in Phoenix, San Francisco, and in all locations combined with the dynamic adjustment. Although the prior study by Chen et al. (2024) demonstrates that the dynamic benchmark adjustment improves alignment between the ADS and benchmark data, because the *Suspected Serious Injury+* benchmark rate is low and the observed collisions is therefore also low, the statistical significance of the result is sensitive to the benchmark. This sensitivity was also true with the data from 7.1 million RO miles presented by Kusano et al. (2024), but those results were confirmed as more miles were accumulated. For this reason, we are presenting the *Suspected Serious Injury+* outcome result as an indication of a result, while the *Any-Injury-Reported* and *Airbag Deployment* results are more certain given the larger number of observed events.

The benchmark adjustments have some effect on the crash type analysis results for the *Any-Injury-Reported* outcome. There are a large number of comparisons performed when decomposing by crash type so this section will not exhaustively describe all the differences. See the attached data for all comparisons. In general without the underreporting and dynamic benchmark adjustments, the outcomes with a smaller reduction tend to not be statistically significant in the absence of the adjustments. The statistically significant *Any-Injury-Reported* Cyclist, Motorcycle, Secondary Crash, and V2V Lateral are no longer statistically significant if neither underreporting or dynamic benchmark adjustment is applied in all locations combined. The Pedestrian, Single Vehicle, and V2V Intersection reductions for *Any-Injury-Reported* outcomes remain statistically significant for all locations combined without underreporting and dynamic adjustments. If the dynamic benchmark adjustment is applied but the underreporting adjustment is not applied, the Cyclist reduction is statistically significant. The *Airbag Deployment* outcome result does not seem to be sensitive to the dynamic benchmark adjustment. The same Single Vehicle and V2V Intersection results were still statistically significant when the dynamic benchmark adjustment was not applied.

Similar trends are found when examining crash type results in Phoenix and San Francisco. Therefore, the *Any-Injury-Reported* reductions in Pedestrian and V2V Intersection and the Single Vehicle and V2V Intersection reductions in the *Airbag Deployment* present the highest confidence results. Overall this supports the conclusions of this study of which crash modes are leading to the observed overall crash rate reductions.

**Trends in Crash Types**

The following sections provide additional information and analyses on select crash types of interest. A complete listing of crashes used in this analysis and their crash type group are provided in the supplemental download file "CSV2 - Crashes with SGO ID and Group Membership 202009-202501-2022benchmark.csv". The data dictionary for this supplemental download can be found here. The columns "Incident Date", "Location Address / Description", and "Zip Code" were removed because they are not pertinent to this study's results. One additional column was added that was not listed in the data dictionary: "Crash Type" indicates which of the 11 crash types examined in this study was assigned to the crash. The supplemental download includes the SGO case identifier, which can be used to look up information about each case in the SGO data published by NHTSA.

**V2V Intersection:** The Waymo RO service had a large, statistically significant reduction in V2V Intersection crashes compared to the human benchmarks across many of the outcomes and locations. Intersection crashes were the single most common crash type in the benchmark, accounting for 57% of *Airbag Deployment* and 39% of *Any-Injury-Reported* crashes in the mileage blended benchmark. There were a total of 4 *Any-Injury-Reported* and 5 *Airbag Deployment* V2V Intersection crashes. In all (100%) of the *Airbag Deployment* and *Any-Injury-Reported* V2V Intersection collisions, the Waymo ADS vehicle was in the responder role, meaning the Waymo ADS needed to react to the unexpected actions of another road user (the initiator). The result of the ADS having a lower crash rate for these outcomes combined with the result that the ADS vehicle was the responder in most V2V Intersection crashes suggests that the ADS vehicle achieves a lower crash rate than the human benchmark by not initiating conflicts that lead to crashes at intersections and responds appropriately to other road users initiating conflicts.

**Cyclists, Motorcyclists, and Pedestrians:** Cyclist, Motorcyclist, and Pedestrian crashes, sometimes called Vulnerable Road User (VRU) crashes, were far more common in the human benchmark for San Francisco than Phoenix and Los Angeles. Table A7 compares the Any-Injury-Reported benchmark crashed vehicle rates and shows that Cyclist crashes occur 20 times more often, Motorcyclist crashes occur 10 times more often, and Pedestrian Crashes occur 16 times more often in San Francisco compared to Phoenix. Cyclist, Motorcyclist, and Pedestrian crashes each occur 8 times more frequently in San Francisco compared to Los Angeles. Therefore, the relative performance of the Waymo ADS can be examined in these crash modes in San Francisco compared to Phoenix and Los Angeles with fewer miles traveled because the benchmark rate in San Francisco is more frequent than in other locations.

Table A8. Comparison of *Any-Injury-Reported* Cyclist, Motorcyclist, and Pedestrian Benchmark Crash Rates (with Dynamic Adjustment) by Location (with ratio of San Francisco to Other Locations in parentheses) (through January 2025, 56.7M RO Miles).

| Location | San Francisco | Phoenix | Los Angeles |
|---|---|---|---|
| Cyclist | 0.80 | 0.04 (20.0) | 0.10 (8.0) |
| Motorcyclist | 0.52 | 0.05 (10.4) | 0.06 (8.7) |
| Pedestrian | 1.12 | 0.07 (16.0) | 0.15 (7.5) |

There were two (2) reported pedestrian crashes during the study period, both of which resulted in an *Any-Injury-Reported* outcome. According to the SGO narrative in case 30270-9459: "A building was located at the intersection, just to the right of the Waymo AV. While the traffic light was green and as the Waymo AV approached the intersection with [XXX], a scooterist who was traveling south on the sidewalk of [XXX] and was previously occluded by the building became visible and entered the intersection against a red light. As the Waymo AV braked for the scooterist, the scooterist made contact with the passenger side of the Waymo AV." In the conflict partner classification used in this study and described by Kusano et al (2023), a person riding a scooter standing in the ambulatory position is classified as a pedestrian. According to the SGO narrative in case 30270-9234: the Waymo vehicle "began to slow to yield to a pedestrian crossing the street ahead from right to left. After the pedestrian had crossed the path of the Waymo AV, the Waymo AV began to accelerate slightly then immediately began to slow again as the pedestrian turned around and approached the Waymo AV. As the Waymo AV was slowing, the pedestrian may have made contact with the driver side of the Waymo AV. … The pedestrian later claimed to have a minor injury." A driver with the mileage blended benchmark crash rate traveling the 56.7M miles that the ADS vehicle traveled would have 23.8 *Any-Injury-Reported* Pedestrian crashes, which leads to the observed statistically significant reduction for the ADS when compared to the mileage blended benchmark and in San Francisco. The ADS reduction in Phoenix and Los Angeles is not yet significant because the benchmark rate is an order of magnitude lower than in San Francisco.

There were two (2) Motorcycle crashes reported for the *Any-Injury-Reported* outcome, both of which were Secondary Crashes. According to the narrative in case 30270-4768: "The Waymo AV was traveling in the leftmost lane on eastbound [XXX], with a group of smaller motorcycles, more commonly known as minibikes, traveling in the right adjacent lane next to and ahead of the Waymo AV. A minibike rider in the group lost control and fell off the minibike a short distance ahead of the Waymo AV in the adjacent lane and the minibike, without a rider, tumbled into the Waymo AV's lane. The Waymo AV applied maximum brake force but made contact with the riderless minibike just before coming to a stop." This crash was classified as a Secondary Crash from the Waymo vehicle's perspective, because the motorcycle without a rider had previously been involved in a loss of control crash involving the ground. The motorcycle rider's reported minor injury was due to the contact with the ground but because the loss of control and subsequent contact with the riderless motorcycle was part of the same crash sequence, this crash is classified as an Any-Injury-Reported outcome. According to the narrative in case 30270-9761: "The Waymo AV was stopped facing East on [XXX] at the intersection with [XXX] at a red light. A motorcycle was stopped behind the Waymo AV, and a pickup truck was stopped behind the motorcycle. A passenger car approached the queue of stopped traffic from behind and made contact with the stopped pickup truck, which was pushed into the motorcycle. The motorcycle then made contact with the Waymo AV." The Waymo vehicle was stopped at a traffic signal when the motorcycle was pushed into the back of the Waymo vehicle by a prior collision. Similar to Pedestrian crashes, an average driver with the benchmark crash rate would have experienced 11.4 Motorcycle crashes over 56.7 million miles, leading to the observed statistically significant reduction in all locations and in San Francisco.

There were three (3) reported cyclist crashes that resulted in a *Any-Injury-Reported* outcome (30270-7075, 30270-8444, and 30270-9015). The first two occurred in San Francisco and the last in Los Angeles. In crash 30270-7075, the ADS vehicle was the responder as it started from a stop at a stop sign and a cyclist turned left from the opposite direction across the path of the ADS vehicle. In crash 30270-8444, "the Waymo AV came to a stop for a traffic stack ahead of the Waymo AV, partially oriented into the dedicated left turn lane at the location where that lane begins, a cyclist merged out of the bike lane and into the general purpose travel lane, and while the cyclist was looking backwards, the front of the bicycle made contact with the rear passenger side of the stopped Waymo AV" according to the SGO case narrative. In both cases, the cyclist was alleged to have sustained an injury. According to the SGO narrative in 30270-9015: the Waymo vehicle "came to a stop for a passenger drop off. A rider in the Waymo AV opened the rear passenger side door as a cyclist was approaching from the rear to the right side of the Waymo AV. The cyclist made contact with the Waymo AVs opened rear passenger side door." Similar to Pedestrian crashes, Cyclist crashes are more frequent in the benchmark in San Francisco compared to Phoenix. There was a statistically significant reduction in Cyclist collision in all locations combined (82% reduction) and in San Francisco (86% reduction). Compared to an average driver with the benchmark crash rate, the Waymo RO service experienced 13.5 fewer *Any-Injury-Reported* Cyclist crashes over 56.7 million miles in all locations combined.

**V2V Opposite Direction:** The benchmark Opposite Direction crashed vehicle rate was low in all locations, accounting for only 2% of the *Any-Injury-Reported* crashes and 3% of Airbag Deployment crashes. The Waymo RO crash rate point estimates were similar in magnitude to and not statistically different from the benchmark rates. In total there were 2 *Any-Injury-Reported* (30270-7157 and 30270-8217) and 2 *Airbag Deployment* (30270-8217 and 30270-8061) V2V Opposite Direction crashes reported by Waymo in the study period. In all crashes Waymo was in the responder role. In 30270-7157, the Waymo ADV "came to a stop to yield to an oncoming vehicle in a narrow roadway with parked vehicles on both sides of the street. As the oncoming SUV passed the Waymo AV, the driver's side rear of the SUV made contact with the driver's side rear of the Waymo AV," according to the SGO case narrative. Review of the onboard sensor data from the Waymo vehicle shows that the collision resulted in minor body damage caused the the rear of the other vehicle scraping the side of the stopped Waymo vehicle. The delta-V in the collision was less than 1 mph, indicating a minor collision. The crash was reported in the SGO as an injury collision because Waymo "received a claim for injuries for three passengers in the" other vehicle, according to the SGO case narrative. Waymo has not received any indication that the crash was *Police-Reported*. Crash 30270-8217 resulted in an airbag deployment and reported injuries. According to the SGO narrative: "a pickup truck traveling in the opposite direction crossed over the double solid yellow line, entering the lane the Waymo AV was occupying. The Waymo AV nudged to the right and applied heavy braking. The pickup truck did not slow down and made contact with the front driver side corner of the Waymo AV. The pickup truck showed evidence of prior crash damage and sparks were emanating from the area of the two front wheels." Lastly, crash 30270-8061 resulted in an airbag deployment. According to the SGO narrative: "As the AV approached the all-way stop sign at [XXX], it detected a vehicle traveling west on [XXX] and crossing into the opposing lane (the lane in which the AV was traveling). The passenger car entered the intersection in the opposing lane without slowing. The Waymo AV slowed down and maneuvered towards the right side of the eastbound lane as the front of the passenger car, still moving in the wrong direction in the eastbound lane, made contact with the front driver side corner of the Waymo AV."

Because the benchmark crash rate is low for V2V Opposite Direction Crashes and there are few Waymo ADS events, it is not possible to draw statistical conclusions about the performance of the Waymo RO service in this crash mode. All observed crashes with these outcomes were responder role scenarios for the Waymo ADS vehicle, suggesting that the Waymo vehicle is not performing unexpected behaviors that led to collisions (i.e., there are no observed crashes where the Waymo vehicle was an initiator).

**V2V Front-to-Rear (F2R):** There were fewer ADS V2V F2R crashes for both the *Airbag Deployment* crashes (43% reduction) and a *Any-Injury-Reported* (23% reduction) when considering all locations combined, neither of which was statistically significant. The comparison for V2V F2R was statistically significant when compared in San Francisco at the *Any-Injury-Reported* outcome level (55% reduction, -83% to -3% confidence interval). The V2V F2R is one of the most common crash modes in the benchmark. A V2V F2R crash involves at least two parties: the striking and struck vehicle. Table A8 compares the ADS and human benchmark crash rates broken out by the striking and struck vehicle assuming 50% of the human benchmark was in the striking and struck vehicle. Note that only the first striking and struck vehicle in a collision sequence are classified as the V2V F2R crash type in this study. Any subsequent vehicles contacted are given the Secondary Crash crash type (e.g., a 3rd vehicle contacted by the initial struck vehicle). The striking and struck designation can be determined from the narrative and data from the ADS crashes. The results show that the ADS vehicle had a statistically significant reduction in F2R Striking for the *Any-Injury-Reported* outcome in Phoenix, San Francisco, and when considering all locations combined and statistically significant reduction in *Airbag Deployment* outcomes when considering Phoenix and all locations combined. There was a statistically significant increase in F2R Struck *Any-Injury-Reported* crashes in Phoenix. All other comparisons were not statistically significant. In general, the Waymo service reduced F2R striking, while not reducing or slightly reducing F2R struck crashes, which resulted in overall V2V F2R rates with similar magnitude to the benchmark rates.

Table A9. Comparison of ADS and Mileage Blended Human Benchmark (Dynamic Benchmark Adjustment) in Any-Injury-Reported and Airbag Deployment F2R Striking and Struck Crashed Vehicle Rates (through January 2025, 56.7M RO Miles).

| Location | Outcome | Crash Type (ADs Vehicle Perspective) | Human IPMM | ADS IPMM | Expected ADS Count Different to Benchmark | ADS to Human Percent Difference | 95% Confidence Interval | |
|---|---|---|---|---|---|---|---|---|
| Phoenix | *Airbag Deployment* | F2R Striking | 0.12 | 0.00 | -3.7 | **-100%*** | **-100%** | **0%** |
| | | F2R Struck | 0.12 | 0.16 | 1.3 | 35% | -63% | 246% |
| | | F2R All | 0.24 | 0.16 | -2.4 | -32% | -82% | 73% |
| | *Any-Injury-Reported* | F2R Striking | 0.22 | 0.00 | -6.9 | **-100%*** | **-100%** | **-47%** |
| | | F2R Struck | 0.22 | 0.48 | 8.1 | **117%*** | **11%** | **281%** |
| | | F2R All | 0.44 | 0.48 | 1.2 | 9% | -44% | 90% |
| San Francisco | *Airbag Deployment* | F2R Striking | 0.10 | 0.00 | -1.9 | -100% | -100% | 103% |
| | | F2R Struck | 0.10 | 0.05 | -0.9 | -46% | -99% | 255% |
| | | F2R All | 0.20 | 0.06 | -2.7 | -73% | -100% | 75% |
| | *Any-Injury-Reported* | F2R Striking | 0.48 | 0.00 | -8.8 | **-100%*** | **-100%** | **-58%** |
| | | F2R Struck | 0.48 | 0.44 | -0.8 | -9% | -66% | 95% |
| | | F2R All | 0.97 | 0.44 | -9.6 | **-55%*** | **-83%** | **-3%** |
| All Locations - Mileage Blended | *Airbag Deployment* | F2R Striking | 0.11 | 0.00 | -6.1 | **-100%*** | **-100%** | **-40%** |
| | | F2R Struck | 0.11 | 0.12 | 0.9 | 14% | -60% | 155% |
| | | F2R All | 0.22 | 0.12 | -5.3 | -43% | -80% | 27% |
| | *Any-Injury-Reported* | F2R Striking | 0.30 | 0.00 | -17.3 | **-100%*** | **-100%** | **-79%** |
| | | F2R Struck | 0.30 | 0.44 | 7.7 | 45% | -12% | 125% |
| | | F2R All | 0.61 | 0.44 | -9.5 | -28% | -56% | 12% |

We performed an additional analysis that classified the ADS vehicle pre-crash movement in F2R crashes to provide additional context for the events. The pre-crash movement was classified into three categories: (a) the ADS vehicle was stopped for more than 5 second prior to the crash, (b) the ADS vehicle was traveling at a constant speed (decelerations less than 0.75 m/s² in the 5 seconds prior to the crash) or accelerating and (c) braking with peak magnitude less than 3.5 m/s² for the 5 seconds prior to a crash. Table A10 shows the pre-crash movement of the ADS vehicle in *Airbag Deployment* and *Any-Injury-Reported* F2R Struck crashes. In 100% of *Airbag Deployment* and 76% of *Any-Injury-Reported* F2R Struck crashes, the ADS vehicle was either stopped, traveling at a constant speed, or decelerating with traffic. The remaining 24% of *Any-Injury-Reported* F2R Struck crashes had braking more than 3.5 m/s² deceleration. Part of Waymo's Safety Framework (Webb et al 2020) includes examination of infield events as one feedback mechanism for future performance improvement. The pre-crash movement analyzed here is insufficient to draw any conclusions about the reasons for the movement or assess if the ADS vehicle's movement behavior contributed to the cause of the F2R struck crashes. The results suggest, however, that a majority of F2R Struck crashes do not involve sudden, high deceleration braking. Third-party liability insurance claims (when an insured vehicle is asked to make an insurance payment by another party to the crash) have been used as a proxy for crash contribution. The most recent third-party liability claims study of the Waymo ADS over 25 million miles of RO driving found that there were no property damage or bodily injury third-party claims for F2R struck. As noted in the discussion section of this paper, there is a need for future research that develops objective models that can be used to quantify crash contribution both in ADS and benchmark data sources. Such a contribution analysis for F2R Struck crashes could consider the following distance and speed of the vehicle behind the ADS vehicle, as well as response times of a typical non-impaired, eyes on conflict (NIEON) model (Engstrom et al 2024). High deceleration braking is required and expected in order to avoid crashes when responding to surprising actions by other road users, which in certain situations that could result in a F2R Struck crashes (e.g., a vehicle suddenly cuts into the ADS vehicle path, requiring the ADS vehicle to brake to avoid a collision). Therefore, a simple kinematic metric is insufficient to determine whether ADS vehicles have an elevated contribution to F2R Struck crashes compared to human drivers. Lastly, although the dynamic benchmark adjustment takes into account the increased driving exposure for the ADS fleet in heavily populated areas which have a higher F2R Struck rate (benchmark of 0.25 IPMM without dynamic benchmark, 0.30 IPMM with dynamic benchmark), the dynamic benchmark may not account for increased exposure to F2R Struck situations during ride-hailing pick-up and drop-offs. It is likely that ride-hailing vehicles spend more time stopped in or near travel lanes due to pick-ups and drop-offs than the overall human driving population, which increases the potential exposure to F2R Struck crashes.

Table A10. ADS Vehicle Pre-Crash Movement in F2R Struck Crashes.

| Outcome | ADS Vehicle Stopped at least 5 seconds | ADS Vehicle Constant Speed or Accelerating | ADS Vehicle Braking < 3.5 m/s² peak deceleration | Any of these 3 Conditions | ADS Vehicle Braking >= 3.5 m/s² | Total F2R Struck |
|---|---|---|---|---|---|---|
| *Airbag Deployment* | 5 (72%) | 1 (14%) | 1 (14%) | 7 (100%) | 0 (0%) | 7 |
| *Any-Injury-Reported* | 5 (20%) | 3 (12%) | 11 (44%) | 19 (76%) | 6 (24%) | 25 |

**Suspected Serious Injury+ Crashes:** Although some were mentioned in the previous crash type descriptions, for completeness this section summarizes the 2 *Suspected Serious Injury+* crashes with police-reported "A" or "K" injuries. The 1 crash with "Serious" maximum severity that occurred during the study period and was reported in the SGO was not included in the *Suspected Serious Injury+* outcome because it had "C" (complaint of pain) maximum police-reported injury severity.. All three crashes had a Secondary Crash crash type.

First, there were two crashes Suspected Serious Injury+ crashes (30270-6579 and 30270-9724). According to the narrative in case 30270-6579: "The Waymo AV was … stopped at a red light at the intersection with [XXX] alongside a passenger vehicle in the right lane. After the light turned green, both the Waymo AV and the adjacent passenger car proceeded into the intersection. While in the intersection, a passenger car traveling west on [XXX] ran the red light and the front left corner and left side of this vehicle made contact with the front right of the Waymo AV and the front of the adjacent passenger car. After impact, the vehicle that ran the red light struck pedestrians that had been standing on the sidewalk on the northwest corner of the intersection." Additional review of the sensor and video data from the Waymo vehicle, it appears the vehicle adjacent to the Waymo vehicle was contacted by the red light running vehicle, then the Waymo vehicle was contacted by the red light running vehicle. Because the Waymo vehicle was involved in the second crash in the collision sequence, this crash was considered a Secondary Crash from the perspective of the Waymo vehicle. A police-report obtained for this crash stated that one of the pedestrians sustained incapacitating injuries (A severity on the KABCO scale). According to the SGO narrative for case 30270-9724: "The Waymo AV, which had no occupants, was traveling northwestbound in the middle of three lanes on [XXX] and came to a stop in a queue of traffic. Shortly after, a passenger vehicle came to a stop behind the Waymo AV. While the Waymo AV and the other passenger vehicle were stopped, an SUV approached from behind at an extreme rate of speed and made contact with the passenger vehicle behind the Waymo AV, which then made contact with the rear bumper of the Waymo AV. … one of the occupants of the vehicles involved in the crash and a domestic animal were declared deceased at the scene." This crash was also a Secondary Crash from the Waymo vehicle's perspective.

One crash (30270-8968) had a maximum SGO-reported severity of "Serious", while a police report obtained for the crash indicated the single injured occupant (not in the Waymo vehicle) had a "C" (complaint of pain) injury. According to the SGO narrative for event 30270-8968: "The Waymo AV came to a stop in a queue of traffic for a red traffic light in the rightmost lane of the two eastbound lanes on [XXX] at the intersection with [XXX]. A passenger car traveling west on [XXX] crossed the double yellow line and made contact with an SUV that was alongside the Waymo AV in the left lane of eastbound [XXX]. The impact caused the passenger side of the SUV to make contact with the driver side of the Waymo AV." The driver of the SUV that was adjacent to the Waymo vehicle was transported to the hospital for medical treatment. This crash was a Secondary Crash crash type because the Waymo vehicle was involved in the second crash in the crash sequence. The NHTSA SGO reporting guidelines define a "Serious" maximum severity as "confirmed or alleged incident where an involved party sustained serious injuries that required hospitalization or emergency treatment." Because this crash had an occupant transported to a hospital by ambulance to receive emergency medical attention, it could be seen as meeting this "Serious" injury requirement. The term "emergency treatment" used in the "Serious" injury definition seems to be a much lower threshold of potential injury (e.g., being transported in an ambulance) compared to "hospitalization," which implies admittance to a hospital as opposed to being treated and released from an emergency department. This lower reporting threshold may lead to difficulties when comparing "Serious" SGO-reported injuries to "A" (incapacitating) police-reported injuries. Waymo has not received any additional information about the nature of these injuries. As the human data used for the *Suspected Serious Injury+* benchmark uses police-reported severity, comparing the Waymo police-reported "K" and "A" severity injuries to the benchmark is the most appropriate comparison.

**Comparisons to Prior Studies**

The current study is an extension of Kusano et al (2024), who performed a retrospective study of the Waymo Driver's overall crash rate in San Francisco and Phoenix over 7.1 million RO miles, but did not perform a crash type analysis due to limited miles. Kusano et al (2024) found a statistically significant reduction in *Police-Reported* (55% reduction, 22% to 77% 95% confidence interval) and *Any-Injury-Reported* (80% reduction, 44% to 96% 95% confidence interval) in all locations combined. The current study found a 65% reduction in *Police-Reported* (57% to 72% 95% confidence interval) and a 79% reduction in *Any-Injury-Reported* (72% to 86% 95% confidence interval) outcomes, which is a similar magnitude with tighter confidence intervals compared to the previous study. Di Lillo et al (2024b) compared the Waymo RO 3rd party liability property damage and bodily injury claims rates to a benchmark of human insured latest-generation vehicles (model years 2018-2021). The study found Waymo RO had an 86% reduction in property damage claims and 90% reduction in bodily injury claims compared to the latest-generation human driven vehicle benchmark. These 3rd party liability reductions are slightly higher than the reductions found in the current study. The 3rd party liability claims results are not directly comparable to the current study results because liability claims are a subset of all crashes where the insured party has some contribution to the cause of the crash. Kusano et al (2024) was led by authors from Waymo, whereas Di Lillo et al (2024b) was led by researchers from Swiss Re.

The examination of crash rates stratified by individual crash types has not yet been performed for data from fully RO ADS. Teoh and Kidd (2017) compared Google Self-driving Car Project (the name of the project prior to the founding of Waymo in 2016) crashes from 2009 through 2015 in Mountain View, CA reported to the California DMV. For this duration, there were no RO operations and all autonomous driving was conducted with an autonomous specialist behind the wheel of the car supervising the autonomy. No underreporting correction was performed for the human crash data. One ADS-to-benchmark comparison done by Teoh and Kid (2017) was *Any-Injury-Reported* crashes. They found the Google project had an overall *Any-Injury-Reported* crash rate of 0.73 IPMM (1 suspected injury reported over 1,372,111 miles), compared to a benchmark derived from police accident reports from the SWITRS database of 3.07 IPMM (without underreporting correction). Teoh and Kidd (2017) also compared the ADS crashes in 6 crash types (single vehicle, rear-end struck, rear-end striking, sideswipe, side impact, and 3+ vehicle), but no statistical testing was possible with the low amount of mileage being evaluated. The single Google *Any-Injury-Reported* crash was a rear-end struck crash (0.73 IPMM), which was not so different from the human benchmark rate of 0.67 IPMM. It is difficult to directly compare the current study to Teoh and Kidd (2017), because the studies had different autonomy types (RO compared to supervised by a human) and different locations (San Francisco, CA and other locations compared to Mountain View, CA). However, in the most comparable outcome (*Any-Injury-Reported*), the current study found similar magnitudes of crash rates as Teoh and Kidd (2017).

Goodall (2021) used all property damage or injury collisions from both testing operations (human supervised) and RO crashes from multiple manufacturers as reported to the CA DMV (form OL316) between October 14, 2014 and March 10, 2020 compared to a benchmark derived from the SHRP-2 NDS in rear-end struck crashes. All Waymo data examined during the study period was under testing operations (not RO), as Waymo did not start RO driving in California until 2022. Like the comparison to Teoh and Kidd (2017), it is difficult to directly compare the results of the current study and Goodall (2021b) because of different autonomy types (rider-only compared to testing operations), different ADS operating locations (San Francisco, CA vs Mountain View, CA), and different benchmark locations and road types (surface streets of San Francisco, CA and other locations compared to multiple collection sites throughout the US including all types of roads). Goodall (2021) found that Waymo vehicles were struck from the rear 2.8 times more often than the human benchmark (10.4 IPMM compared to 3.6 IPMM). The current study only examines RO crashes and miles, much of which have been collected after the March 2020 cutoff used by Goodall (2021). The SHRP-2 benchmark used by Goodall (3.6 IPMM) is most comparable to the *Any Property Damage or Injury* benchmark from Scanlon et al. (2024a), which applies an underreporting correction to police-reported data. The *Any Property Damage or Injury* V2V F2R Struck mileage blended benchmark for the current study period was 1.09 IPMM, which was lower than the SHRP-2 benchmark.

As noted in Scanlon et al. (2024a), benchmark crash rates can be influenced by factors such as location and road type, so it is unclear how applicable the SHRP-2 benchmarks are to the locations where Waymo RO operations are. Nevertheless, the Waymo RO *Any Property Damage or Injury* crash rate for F2R Struck was 2.54 IPMM considering all locations combined. The human benchmark from Goodall (2021) is 1.4 times greater than the Waymo rate in this outcome and crash mode, which is the opposite conclusion Goodall (2021) drew using the early Waymo data. As stated in Scanlon et al (2024a), the geographic region appears to have a large influence on benchmark crash rate, so it is unlikely the sample of test locations across the US used in SHRP-2 is representative of the urban areas Waymo currently operates in. The Waymo crash rate (2.54 IPMM) is 2.3 times the *Any Property Damage or Injury* V2V F2R Struck mileage blended benchmark (1.09 IPMM) using police-reported data with the underreporting correction from Blincoe et al (2023). As noted by Kusano et al. (2024), it is difficult to draw conclusions about the most broad *Any Property Damage or Injury* outcome because of uncertain underreporting and inclusion criteria in the human benchmark. Additionally, there is documented underreporting in property damage only in California crash data, as police jurisdictions are not required by law to report property damage only crashes in the aggregate state crash data that is publicly available (Scanlon et al 2024a). The confidence intervals of the *Airbag Deployment and Any-Injury-Reported* F2R struck comparisons done in the current study are still quite wide in F2R Struck crashes, so it may be too soon to draw conclusions. However, these higher severity outcomes are less sensitive to underreporting or inclusion criteria than the *Any Property Damage or Injury* outcome level.

Chen and Shladover (2024) compared ADS crashes reported as part of the NHTSA SGO in San Francisco to several human benchmarks. The crash benchmarks highlighted in Chen and Shladover (2024) were self-reported crash events reported by Uber as part of the regulatory filings made to the California Public Utilities Commission (CPUC) as part of Transportation Network Company (TNC) reporting and two naturalistic crash rates from a study performed by the University of Michigan Transportation Research Institute (UMTRI) and the Virginia Tech Transportation Institute (VTTI) which used both rental vehicles primarily used as ride-hailing vehicles and instrumented vehicles. The CPUC benchmark was 15.5 IPMM and the UMTRI/VTTI study reported two estimated rates of 36.2 IPMM and 50.5 IPMM. These human benchmarks were compared to all Waymo RO collisions reported as part of the NHTSA SGO between September 2022 and August 2023, resulting in a crash rate of 14.1 IPMM over 994,842 RO miles (n=14 crashes). The NHTSA SGO crash rate represents an *Any Property Damage or Injury* crash rate, as ADS manufacturers are required to report a crash with any amount of property damage. The CPUC TNC crash data are self-reported by the company, which requires the TNC to be notified about the collision. One of the primary challenges with using CPUC TNC crash data, that is also a limitation noted by Chen and Shladover (2024), is the lack of any documented reporting thresholds (see RAVE Checklist recommendation 1A in Scanlon et al., 2024). Seemingly, inclusion in this dataset requires (1) some contact event occurs that causes property damage or injury, (2) Uber is notified of that event having occurred, and (3) the event is deemed sufficient for reporting to CPUC TNC. Without clear guidance on the protocol for documenting each event, it is impossible to know the amount of potential missingness present in the data. Chen and Shladover (2024) reported that there was high missingness of reported fields in newer data, which required the authors to use only data from Uber from calendar year 2020. It's likely that higher severity collisions that result in substantial property damage or injury are reported by either a rider, driver, or can be automatically detected through other means. Minor collisions, which make up the majority of collisions, may not be reported, as in other human crash databases. Drivers may have an incentive to not report minor collisions, as they may suspect that reported collisions may impact their driver rating or their ability to meet the requirements to remain as a driver on the TNC network. For this reason of underreporting, the UMTRI/VTTI study used a combination of instrumented vehicles with manual review of sensor data to find most contact events with a larger fleet of telematics-connected rental vehicles to estimate this underreporting of minor collisions. Therefore, without further study into the reporting practices and underreporting amount in the TNC crash data, it does not seem appropriate to compare ADS crash rates from the NHTSA SGO to the CPUC TNC crash data. Chen and Shaldover (2024) used Driverless Pilot program mileage reported by ADS ride-hailing operators to form crash rates for Waymo's RO service, as the study was performed before Waymo started to self-publish RO miles. Using the data from the current study, there were 14 NHTSA SGO crashes in San Francisco between September 2022 and August 2023 (i.e., through the end of July 2023) with 1.004 million RO miles (1% difference to Chen and Shaldover, 2024).

Cummings and Bauchwitz (2024) compute several metrics related to current ADS, including a comparison of crash rates to human benchmarks (see Cummings and Bauchwitz Figure 3). The Waymo crashes were sourced from the NHTSA SGO and include a mix of RO (SAE level 4 without a driver present) and supervised autonomy (SAE level 4 with a autonomous specialist behind the wheel supervising the autonomy) between January 2022 and December 2023 resulting in 256 Waymo collisions. Using the current study's data, there were 37 Waymo RO crashes during this time period with the remaining where the autonomy was supervised by a human behind the wheel. As noted in the current study, RO driving is the most representative of the performance of the ADS in its intended operating mode. There were two human benchmarks used by Cummings and Bauchwitz (2024): first, a similar TNC benchmark from CPUC data as used by Chen and Shladover (2024) and second, national police-reported crash and VMT data. As discussed above, it is unclear what the reporting threshold is and how much underreporting there is in the CPUC TNC data. There is assumed to be little underreporting in the SGO data and no lower limit for reporting threshold. Therefore, without supporting data that quantifies the lower reporting threshold and/or underreporting in the CPUC TNC data, it is not appropriate to draw the conclusion that Waymo has a similar crash rate than TNC drivers in San Francisco. Furthermore, it is inappropriate to compare a human police-reported crash rate to the overall NHTSA SGO crash rate because police-reported crashes require a certain level of property damage or injury to be reported. Numerous previous studies have noted this limitation and cautioned against comparing ADS data like the NHTSA SGO directly to human police-reported data (Teoh and Kidd 2017, Scanlon et al 2024, Chen and Shladover 2024). The current study focuses on higher severity outcomes (*Any-Injury-Reported*, *Airbag Deployment*, and *Suspected Serious Injury+*), as these injury-relevant crashes have been the traditional focus of traffic safety. Due to vertical heterogeneity (i.e., that the causes and incidence rates of serious collisions may be different and/or uncorrelated to more severe collisions) (Knipling 2017), performance in lower severity crashes like the *Any Property Damage or Injury* outcome represented by all SGO crashes should not be extrapolated to make determinations about more severe crash outcomes.

**Appendix References**

**Study Conformance to the RAVE Checklist**

The following table is a self-assessment (by the authors) on this study's conformance to the RAVE Checklist. Overall, we believe our study conforms to the recommendations outlined in the RAVE checklist and serves as a useful, unbiased indicator of safety impact. There are areas where conformance is not fully satisfactory, which we attribute primarily to limited human benchmarking data quality that restricts the ability to fully account for potential confounders. These identified areas serve as a useful indicator for future research areas to improve upon.

Table A11. RAVE Checklist Evaluation for Quality and Validity Recommendations.

| # | Recommendation | Recommendation Type | Actions | Conformance Level | Justification |
|---|---|---|---|---|---|
| 1a. | Reporting differences were considered and addressed, if necessary, through methodological choices. | Required | <ul><li>ADS NHTSA SGO and human benchmark police report data have different data labels specifying injury outcomes (*any-injury-reported*, *suspected serious injury+*) and *airbag deployment*.</li><li>Data fields were relied upon in both data sources that represent equivalent severity levels.</li><li>For potential ADS *suspected serious injury+* events, the police report was obtained to determine whether a "K" or "A" level occupant injury occurred (based on the specific "A" level injury definition for that geographic area).</li><li>Geographic-specific benchmark subselection routines were used to best match the corresponding SGO outcomes.</li><li>National underreporting estimates from NHTSA were used to account for underreporting in the *any-injury-reported* benchmark only.</li></ul> | Satisfactory | <ul><li>Perfect alignment between SGO and human crash data is limited due to the available data fields.</li><li>For each of the reported outcomes, the selected methodological choices enabled an unbiased comparison between the benchmark and ADS driving groups.</li></ul> |

| # | Recommendation | Recommendation Type | Actions | Conformance Level | Justification |
|---|---|---|---|---|---|
| 1b. | Exposure differences were considered and addressed, if necessary, through methodological choices. | Recommended | • Benchmarks were subset to include similar conditions as the ADS ODD: (a) same geographic location (county), (b) restricted to surface streets, (c) involving passenger vehicles.<br>• Benchmarks were adjusted proportional to the amount of driving done by the ADS fleet (dynamic benchmark adjustment described by Chen et al 2024). | Somewhat | • Several established confounders were accounted for with their known effects being described.<br>• There remains unaccounted for confounders due to data and methodology availability.<br>• More research and data is needed to further understand the effect of additional confounders. |
| 1c. | Outcome and exposure units were matched between ADS and benchmark. | Required | • ADS and benchmark crash rates are expressed in a crashed vehicle rate (i.e., number of crashed vehicles divided by the total mileage driven by the vehicle population).<br>• ADS data use the number of ADS vehicles involved in a crash divided by the total miles traveled by all ADS vehicles.<br>• Human benchmarks use the number of crashed vehicles divided by the population VMT on an annual basis. | Satisfactory | • Established best practices are being utilized. |
| 2a. | The methodological choices were selected to minimize bias and, when necessary, favored conservatism. | Required | • Many data alignment shortcomings have been identified and addressed by careful creation of a subselection routine, an *any-injury-reported* underreporting adjustment, and dynamic benchmark adjustment.<br>• Crashes involving unknown vehicle type in the benchmark were assumed to follow the same distribution of vehicle type as in | Satisfactory | • No clear bias is evident from the applied methodology.<br>• Human crash and mileage data have limited data fields for aligning upon. The custom geographic-specific routines relied upon for aligning driving exposure are generally limited by this data quality. |

| # | Recommendation | Recommendation Type | Actions | Conformance Level | Justification |
|---|---|---|---|---|---|
| | | | the overall population, thus accurately but conservatively counting vehicles in the benchmark.<br>• A national underreporting estimate provided by NHTSA was used to estimate the *any-injury-reported* crash rates. | | The absolute accuracy of the crash and mileage subselection and dynamic benchmark adjustment are relatively unknown.<br>• In the absence of geographic specific underreporting estimates, NHTSA's national underreporting estimates for any-injury-reported represent a best estimate and do not introduce any clear bias to the analysis. |
| 2b. | Sensitivity analyses, if applicable, were used to explore the effect of methodological decisions on results. | Recommended | • Sensitivity analyses were performed on the effect of underreporting adjustment used in the any-injury-reported benchmark<br>• Sensitivity analyses were performed on the effect of dynamic benchmark adjustment in the benchmark | Satisfactory | • The sensitivity analyses provided needed context regarding the effect of influential modeling decisions on the results. |
| 3a. | Measurable outcomes, if used, were selected with consideration and discussion of potential biases. | Recommended | • Outcomes were directly measurable in both ADS and benchmark data sources (*any-injury-reported, airbag deployment, suspected serious injury+*)<br>• Potential bias in underreporting was discussed. | Satisfactory | • Potential biases were identified and discussed for the analyzed outcomes.<br>• The range of potential outcomes were selected based on prior established research. Excluded outcomes were discussed for their potential for introducing bias.<br>• For any airbag deployment, the Waymo vehicle's airbag will deploy in all crash |

| # | Recommendation | Recommendation Type | Actions | Conformance Level | Justification |
|---|---|---|---|---|---|
| | | | | | configurations regardless of occupancy, which makes it an unbiased outcome level.<br>• The *any-injury-reported* and *suspected serious injury+* benchmarks are affected by occupancy (e.g., presence, position, demographics, belt status). The decision to not include confounder controls for occupancy are described and do not bias the results with respect to the stated research questions. |
| 4a | Transformation methodologies, if used, were well documented, anchored in scientific literature, and appropriate for the evaluation scope. | Required | • The underreporting adjustment used for the any-injury-reported benchmark is based on established literature (Blincoe et al 2023) and has been used in past ADS safety impact research (Kusano et al 2024). | Satisfactory | • The underreporting correction methodology was based on established literature and methodology, and represents a best available estimate for *any-injury-reported* crash rates. |
| 4b | Potential biases and sources of uncertainty introduced by transformation methodologies, if used, were addressed. | Recommended | • A wide array of outcomes was examined, some of which do not have underreporting adjustments. This gives multiple signals that the conclusions of the study are not dependent on the underreporting corrections.<br>• See also sensitivity analyses of underreporting and dynamic benchmark adjustment (#2b). | Somewhat | • In the absence of geographic specific underreporting estimates, NHTSA's national underreporting estimates for *any-injury-reported* represent a best estimate and do not introduce any clear bias to the analysis.<br>• More work is needed to validate the use of a national |

| # | Recommendation | Recommendation Type | Actions | Conformance Level | Justification |
|---|---|---|---|---|---|
| | | | | | estimate and/or remove reliance on underreporting estimates. |
| 5a. | Major sources of uncertainty were identified, discussed, and/or accounted for in the analysis. | Required | • Subsetting data based on comparable outcomes and influential factors addressed many potential sources of bias. <br> • See also sensitivity analyses of underreporting and dynamic benchmark adjustment (#2b). | Satisfactory | • When possible, sources of statistical uncertainty were directly incorporated into the analysis. <br> • In cases where statistical uncertainty was impractical (underreporting and dynamic benchmark adjustment), sensitivity analyses were relied upon. <br> • Several less influential sources of uncertainty were identified, discussed, and remain: <br> ○ Proportion of unknown vehicle types estimated to be passenger vehicles <br> ○ Uncertainty in the total human VMT due to sampling strategy. <br> ○ Degree to which miscodings and input errors are creating bias and uncertainty in the estimates. |
| 5b. | Statistical conclusions, if drawn, were reported following reasonable | Required | • Rate ratio confidence intervals based on a Poisson model (Nelson 1970) were used to evaluate statistically significant | Satisfactory | • Established statistical methodology was relied upon. |

| # | Recommendation | Recommendation Type | Actions | Conformance Level | Justification |
|---|---|---|---|---|---|
| | statistical testing. | | differences between the benchmark and ADS crash rates<br>● Analysis in past studies has shown that the Nelson 1970 method produces wider (i.e., more conservative) confidence intervals compared to parametric bootstrapping (Kusano et al 2024) | | |

Table A12. RAVE Checklist Evaluation for Transparency Recommendations.

| # | Recommendation | Recommendation Type | Actions | Conformance Level | Justification |
|---|---|---|---|---|---|
| 6a | The names of all data sources relied upon were specified. | Required | ● Data sources were named | Satisfactory | ● Sufficient information was provided. |
| 6b | The origins of the sources were specified. | Required | ● The data sources origins were stated. | Satisfactory | ● Sufficient information was provided. |
| 6c | The date ranges of data were specified. | Required | ● Data set ranges were specified | Satisfactory | ● Sufficient information was provided. |
| 6d | The data reporting frequencies (e.g., annual vs. monthly) were specified. | Required | ● Reporting frequencies were specified | Satisfactory | ● Sufficient information was provided. |
| 6e | The data sampling scheme (if applicable) was specified. | Required | ● SGO sampling scheme (i.e., all RO collisions, excluding testing operations) was specified.<br>● The police-reported sampling schemes were described.<br>● The Waymo mileage recording methodology was stated.<br>● The human benchmark mileage sampling methodologies were described. | Satisfactory | ● Sufficient information was provided. |

| # | Recommendation | Recommendation Type | Actions | Conformance Level | Justification |
|---|---|---|---|---|---|
| 6f | References to any relevant documentation detailing the data source (e.g., data dictionaries) were included. | Required | • Documentation data are cited and readily available from public entities providing data.<br>• References are made to the Waymo Safety Impact Data Hub that contains data information. | Satisfactory | • Data documentation is all readily available and cited. |
| 6g | Data features influencing inclusion criteria and data reporting were specified. | Recommended | • Inclusion criteria were specified with data source specific reporting considerations being described. | Satisfactory | • Sufficient information was provided. |
| 6h | Other peculiarities about the data were noted that may have influenced study results. | Recommended | • Not applicable. | Satisfactory | • The components outlined in 6A-6G cover the range of data |
| 7a | Descriptive statistics were presented showing differences in ADS and benchmark sources. | Recommended | • Not applicable. | Somewhat | • |
| 7b | A table was used to showcase differences between the ADS and benchmark data sources. | Recommended | • Table of mileage by year and location was shown. | Somewhat | • |
| 8a | The ADS systems being evaluated were specified. | Recommended | • The ADS system was specified | Satisfactory | • Sufficient information was provided. |
| 8b | The driving locations (e.g., road type and geographic areas) were specified. | Required | • Road type and geographic locations were specified | Satisfactory | • Sufficient information was provided. |
| 8c | Presence of a human operator or supervisor were specified. | Required | • Only RO data was studied, and this was stated | Satisfactory | • Sufficient information was provided. |
| 8d | Other relevant features (if applicable) about the driving environment were specified. | Recommended | • The ADS driving environment was specified in the methods section.<br>• The Waymo RO driving environment is | Satisfactory | • Sufficient information was provided. |

| # | Recommendation | Recommendation Type | Actions | Conformance Level | Justification |
|---|---|---|---|---|---|
| | | | the same as described by Kusano et al (2024), which is referenced in this study. | | |
| 9a | The methodology was described with enough detail to enable replication. | Required | • Every step relied upon for replicating the current study was provided within the publication. | Satisfactory | • The documentation was thorough enough for enabling replication. |
| 9b | Any relied upon published methodology, if applicable, was described and referenced. | Required | • The methodology relied on peer-reviewed studies (Chen et al 2024, Kusano et al 2024, Scanlon et al 2024, Nelson 1970) | Satisfactory | • All relied upon previously published methodology was described and referenced. |
| 9c | Any additional data annotations or classifications on top of the raw data were described. | Required | • ADS data annotations were described and provided (outcome coding and manual review, crash type coding) | Satisfactory | • A full accounting of custom data annotations were provided. |
| 10a | Potential biases of data source limitations were presented and discussed. | Required | • Missing property damage crash reporting in the California SWITRS database was discussed, which can lead to underestimates of the human *airbag deployment* rate estimates.<br>• Injury reporting in the NHTSA SGO and police-reported benchmark data was discussed | Satisfactory | • Known potential sources of bias surrounding the data sources were discussed. |
| 10b | Potential biases of analytical decisions were presented and discussed. | Required | • See data quality recommendations (1a, 1b, 2a, 2b, 3a, 4a, 4b, 5a, 5b) | Satisfactory | • Analytical decision potential biases are covered extensively in the noted sections. |
| 10c | Limitations around assessment scope were presented and discussed. | Required | • Limitations around crash contribution were discussed<br>• Limitations for higher-severity outcomes were discussed | Satisfactory | • The noted limitations around assessment scope properly contextualize the current assessment scope. |
| 10d | Effects of the cumulative | Recommended | • Interpretation of the results discussed | Somewhat | • The methodology is noted to |

| # | Recommendation | Recommendation Type | Actions | Conformance Level | Justification |
|---|---|---|---|---|---|
| | limitations on results interpretations were presented and discussed. | | | | generally be expected to produce an unbiased estimate.<br>• The sensitivity analyses provide a means for investigating the magnitude of certain modeling decisions.<br>• Relative to existing studies, the current effort is noted to extend the state-of-the-art. |

Table A13. RAVE Checklist Evaluation for Interpretation Recommendations.

| # | Recommendation | Recommendation Type | Actions | Conformance Level | Justification |
|---|---|---|---|---|---|
| 11a | Relevant literature motivating the study was presented and discussed. | Required | • Past ADS studies and motivation listed in the introduction study.<br>• The ways in which the current study extends this body of work - from a research scope and need perspective - is extensively laid out.<br>• The analyzed outcomes were preselected from prior studies to enable transparent selection of outcome focus. | Satisfactory | • The actions implemented sufficiently outline the motivation for this work with respect to the existing literature.<br>• All known literature on the topic is noted. |
| 11b | Relevant literature related to or influencing the study design choices were presented and discussed. | Required | • The current study builds upon and extends an extensive body of literature.<br>• All of this literature built upon is presented and discussed. | Satisfactory | • All of the literature built upon is presented and discussed.<br>• All known literature on the topic is noted. |
| 11c | The results and conclusions of the study were presented | Required | • Comparison to past research was done in the discussion section | Satisfactory | • The current study results are presented alongside the |

| # | Recommendation | Recommendation Type | Actions | Conformance Level | Justification |
|---|---|---|---|---|---|
| | alongside relevant literature. | | | | entire body of literature on RO ADS safety performance assessment.<br>• Similarities and differences between those studies were sufficiently described. |
| 11d | The justification for excluded outcome measures and analytical lenses were explained. | Recommended | • Justification for higher-severity outcomes was presented in the introduction section. | Satisfactory | • The study described the full set of potential outcome measures considered that were based on prior research.<br>• The selected outcomes were then justified according to their relevancy for measuring statistical signficance at the current deployment scale. |
| 12a | Research questions were stated. | Required | • The research questions were stated in the introduction section. | Satisfactory | • Sufficient information was provided. |
| 12b | Research questions were specific and appropriately scoped according to the study design. | Required | • The research question was scoped. | Satisfactory | • Sufficient information was provided. |
| 13a | The conclusions logically follow the stated research questions. | Required | • The conclusions answer the stated request questions. | Satisfactory | • Sufficient information was provided. |
| 13b | The conclusions stated were restrained to only what could be logically inferred given the study design. | Recommended | • Conclusions were restricted to the research question. | Satisfactory | • Sufficient information was provided. |
| 13c | The conclusions were | Recommended | • Limitations were discussed and the | | • The study uses |

| # | Recommendation | Recommendation Type | Actions | Conformance Level | Justification |
|---|---|---|---|---|---|
| | appropriate given the study's limitations. | | cumulative effects of those limitations are not expected to create alternative statistical findings and study conclusions. (See also #10d) | | • state-of-the-art data sources techniques to align the benchmarks<br>• Sensitivity analysis show the results of the study are insensitive to adjustments. |
| 14a | Contributing factors were accounted for and discussed in methodology relating findings from lower to higher severity outcomes (if applicable). | Required | • Lower severity outcomes were not used to extrapolate higher-severity performance.<br>• Disaggregated by crash type allows for grouping of contributing factors common to those crash types. | N/A | • Not applicable. |
| 14b | Limitations around any methodology relating findings from lower to higher severity outcomes were discussed (if applicable). | Required | • Not applicable. | N/A | • Not applicable. |
| 15a. | Rates were presented in terms of incidents per exposure units. | Recommended | • Rates were presented in incidents per million miles. | Satisfactory | • This recommended reporting practice was followed. |